\documentclass[10pt,twocolumn,letterpaper]{article}

\usepackage[pagenumbers]{cvpr} % To force page numbers, e.g. for an arXiv version

\usepackage{graphicx}
\usepackage{amsmath}
\usepackage{amssymb}
\usepackage{booktabs}
\usepackage{color}
\usepackage[accsupp]{axessibility}  % Improves PDF readability for those with disabilities.

\usepackage[pagebackref,breaklinks,colorlinks]{hyperref}
\usepackage{cvpr}      % To produce the REVIEW version
\usepackage{graphicx}
\usepackage{amsmath}
\usepackage{amssymb}
\usepackage{booktabs}
\usepackage{color}
\usepackage{multicol}
\usepackage{multirow}
\usepackage{array}

\usepackage[capitalize]{cleveref}
\crefname{section}{Sec.}{Secs.}
\Crefname{section}{Section}{Sections}
\Crefname{table}{Table}{Tables}
\crefname{table}{Tab.}{Tabs.}

\def\cvprPaperID{03324} % *** Enter the CVPR Paper ID here
\def\confName{CVPR}
\def\confYear{2023}

\definecolor{wrq_color}{RGB}{16, 172, 132}
\definecolor{rewriting_color}{RGB}{120, 90, 15}

\begin{document}

\title{RIDCP: Revitalizing  Real Image Dehazing via High-Quality Codebook Priors}

\author{
Rui-Qi Wu\textsuperscript{1}\quad  Zheng-Peng Duan\textsuperscript{1}\quad Chun-Le Guo\textsuperscript{1}\thanks{Corresponding author}\quad Zhi Chai\textsuperscript{2}\quad Chongyi Li\textsuperscript{3}\\
\textsuperscript{1}VCIP, CS, Nankai University\quad
\textsuperscript{2}Hisilicon Technologies Co. Ltd.\\
\textsuperscript{3}S-Lab, Nanyang Technological University\\
{\tt\small \{wuruiqi, adamduan0211\}@mail.nankai.edu.cn, guochunle@nankai.edu.cn,}\\
{\tt\small chaizhi2@huawei.com, chongyi.li@ntu.edu.sg}\\
{\tt\small \href{https://github.com/RQ-Wu/RIDCP_dehazing}{[Code]}\quad\href{https://rq-wu.github.io/projects/RIDCP}{[Website]}}
}

\maketitle

%%%%%%%%% ABSTRACT
\begin{abstract}
% In this work, we  present a new paradigm, for real image dehazing.
% %
% Our key discoveries are (1) a phenomenological degradation pipeline can reduce the gap between existing  hazy image synthesis and real cases; (2) a discrete codebook prior, pre-trained on high-quality images, is surprisingly effective for image dehazing.
% %
% Instead of adopting the de facto physical scattering model, we rethink the degradation  of real hazy images and propose a phenomenological pipeline, which facilitates neural networks for real hazy scenes.
% %
% Inspired by the observation that a pre-trained codebook  has the capability of mapping a hazy image to its clean version, we leverage the high-quality codebook prior for image dehazing.
% %
% To achieve that, we propose a  feature re-matching to mitigate the domain gap between high-quality training data and low-quality hazy input in the code prediction.
% %
% Besides, we design a normalized feature modulation to complement the information loss caused by vector-quantized operation.
% %via introducing the features of the continuous space.
% %
% %These endow optimal dehazing performance. 
% %
% %Our method is also appealing in its controllability for dehazing levels.
% %
% Extensive experiments verify our discoveries and show the superior performance of our method in real image dehazing. 
% %
% Code and data will be released.

Existing dehazing approaches struggle to process real-world hazy images
owing to the lack of paired real data and robust priors.
In this work, we present a new paradigm for real image dehazing from the perspectives of synthesizing more realistic hazy data and introducing more robust priors into the network.
Specifically, (1) instead of adopting the de facto physical scattering model, we rethink the degradation  of real hazy images and propose a phenomenological pipeline considering diverse degradation types.
(2) We propose a \textbf{R}eal \textbf{I}mage \textbf{D}ehazing network via high-quality \textbf{C}odebook \textbf{P}riors (RIDCP).
Firstly, a VQGAN is pre-trained on a large-scale high-quality dataset to obtain the discrete codebook, encapsulating high-quality priors (HQPs).
After replacing the negative effects brought by haze with HQPs, the decoder equipped with a novel normalized feature alignment module can effectively utilize high-quality features and produce clean results.
However, although our degradation pipeline drastically mitigates the domain gap between synthetic and real data, it is still intractable to avoid it, which challenges HQPs matching in the wild. 
%
% Thus, we propose a controllable HQPs matching (CHM) to help the network find better HQPs.
Thus, we re-calculate the distance when matching the features to the HQPs by a controllable matching operation, which facilitates finding better counterparts.
%
% Users are allowed to control the degree of enhancement flexibly by CHM, and a recommended adjustment is also given by an explainable solution.
% A recommended adjustment is given by an explainable solution, and users are also allowed to control the degree of enhancement flexibly follow their perceptions. 
    We provide a recommendation to control the matching based on an explainable solution. Users can also flexibly adjust the enhancement degree as per their preference.
Extensive experiments verify the effectiveness of our data synthesis pipeline and the superior performance of RIDCP in real image dehazing.
\end{abstract}

\vspace{-0.6cm}
\section{Introduction}
\label{sec:intro}

% \begin{figure}[!t]
%     \centering
%     \begin{subfigure}{.48\linewidth}
%         \includegraphics[width=\linewidth, height=2.7cm]{figures/intro/input.png}
%         \put(-85, -10){\small{(a)~Hazy input}}
%         \vspace{0.05cm}
%     \end{subfigure}
%     \begin{subfigure}{.48\linewidth}
%         \includegraphics[width=\linewidth, height=2.7cm]{figures/intro/colorline.png}
%         \put(-90, -10){\small{(b)~Color Line~\cite{fattal2014dehazing}}}
%         \vspace{0.05cm}
%     \end{subfigure}
%     \begin{subfigure}{.48\linewidth}
%         \includegraphics[width=\linewidth, height=2.7cm]{figures/intro/PSD.png}
%         \put(-80, -10){\small{(c)~PSD~\cite{chen2021psd}}}
%         \vspace{0.05cm}
%     \end{subfigure}
%     \begin{subfigure}{.48\linewidth}
%         \includegraphics[width=\linewidth, height=2.7cm]{figures/intro/ours.png}
%         \put(-70, -10){\small{(d)~Ours}}
%         \vspace{0.05cm}
%     \end{subfigure}
%     \caption{Visual comparisons on a typical hazy image.
%     As seen from the figure,
%     the proposed method generate more visually pleasing result than other two state-of-the-art image dehazing approaches. \lichongyi{Please follow the introduction to replace the results with those of more diverse methods. Also, we may need to select a more attractive example.}
%     }
%     \vspace{-0.7cm}
%     \label{fig:intro_compare}
% \end{figure}
\begin{figure}
    \centering
    \includegraphics[width=\linewidth]{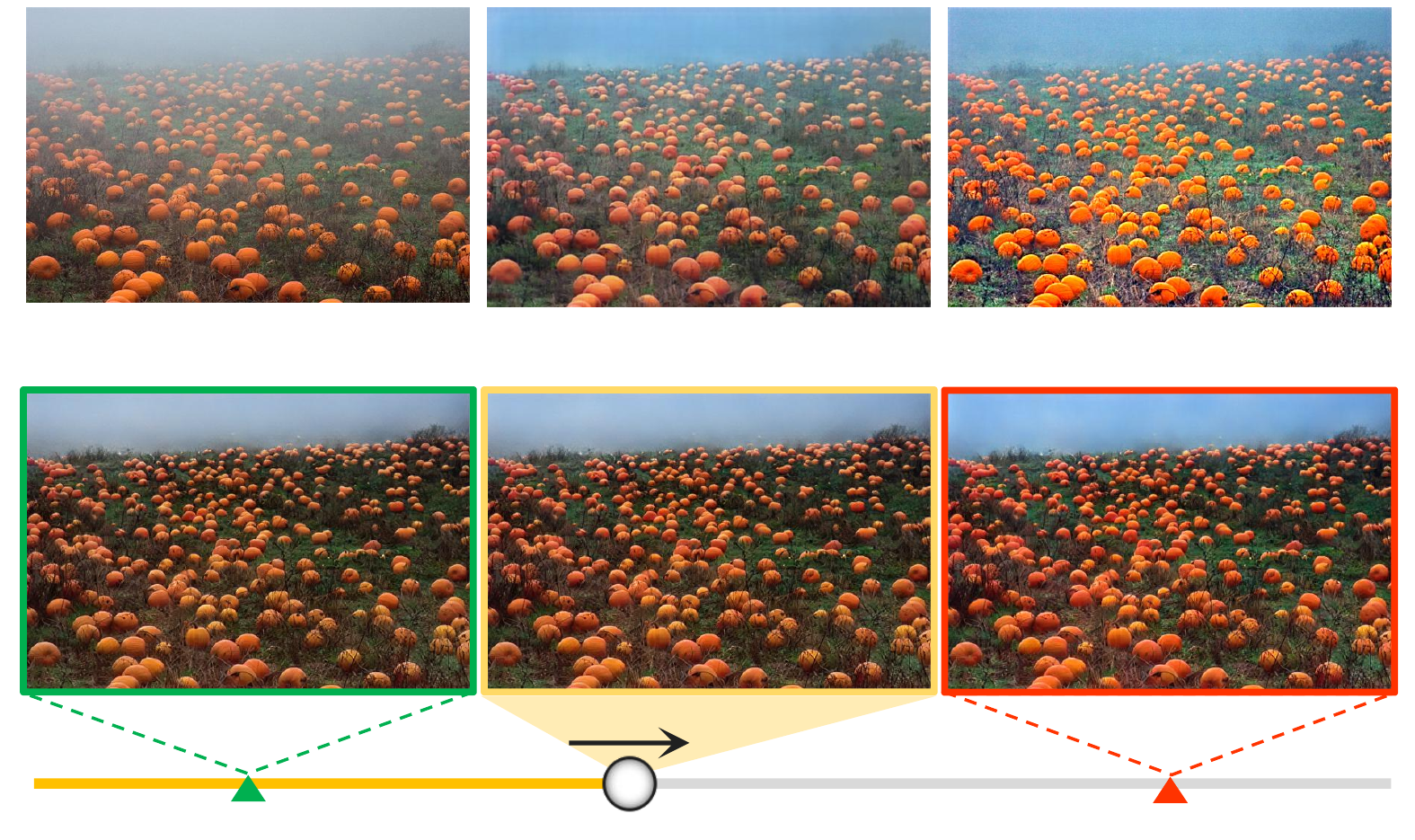}
    \put(-221, 80){\small{(a)~Hazy input}}
    \put(-143, 80){\small{(b)~DAD~\cite{shao2020domain}}}
    \put(-62, 80){\small{(c)~PSD~\cite{chen2021psd}}}
    \put(-136, -8){\small{(d)~Ours}}
    % \put(-153, 15){\footnotesize{\textit{recommended}}}
    % \put(-128, 5.3){$\longrightarrow$}
    \vspace{-0.25cm}
    \caption{Visual comparisons on a typical hazy image. The proposed method generates cleaner results than other two state-of-the-art real image dehazing approaches.  The enhancement degree of our result can be flexibly adjusted  by adopting different parameters in the real-domain adaptation phase. The image with a golden border is the result obtained under our recommended parameter.}
    \vspace{-0.5cm}
    \label{fig:intro_compare}
\end{figure}
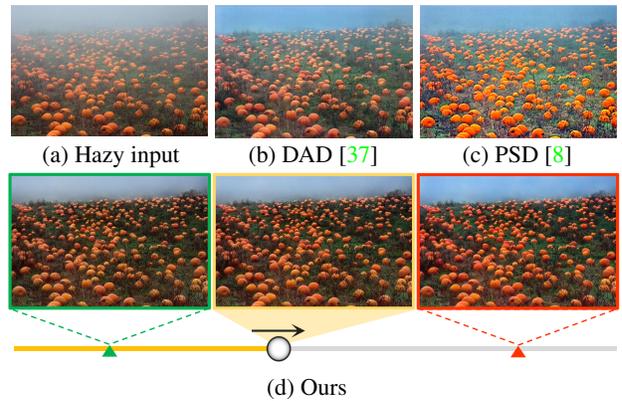

%%%%%%%%%%%%%%%%%%%%%%%%%%%%%%%%%%%%%%%%%%%%%%%%%%%%%%%%%%%%%%%%
Image dehazing aims to recover clean images from their hazy counterparts, which is essential for computational photography 
and high-level tasks~\cite{huang2020dsnet, sakaridis2018model}.
The hazy image formulation is commonly described by a physical scattering model:
\begin{equation}
I(x) = J(x)t(x) + A(1-t(x)),
\label{eq:physical_model}
\end{equation}
where $I(x)$ denotes the hazy image and $J(x)$ is its corresponding clean image. The variables $A$ and $t(x)$ are the global atmosphere light and transmission map, respectively.
The transmission map $t(x) = e^{\beta d(x)}$ depends on scene depth $d(x)$ and haze density coefficient $\beta$.

Given a hazy image, restoring its clean version is highly ill-posed.
%
%However, estimating haze-free image from a hazy input is challenging since it is an ill-posed problem.
%
To mitigate the ill-posedness of this problem,  various priors, e.g., dark channel prior~\cite{he2010single}, color attenuation prior~\cite{zhu2014single}, and color lines~\cite{fattal2014dehazing} have been proposed in existing traditional methods. 
Nevertheless, the statistical priors cannot cover diverse cases in real-world scenes, leading to suboptimal dehazing performance.

With the advent of deep learning, image dehazing has achieved remarkable progress.
Existing methods either adopt deep networks to estimate physical parameters~\cite{cai2016dehazenet, ren2016single, li2017aod} or directly restore haze-free images~\cite{liu2019griddehazenet, qin2020ffa, dong2020multi, guo2022image, ye2021perceiving}.
    However, image dehazing neural networks perform limited generalization to real scenes, owing to  the difficulty in \textit{collecting large-scale yet perfectly aligned paired training data} and \textit{solving the uncertainty of the ill-posed problem without robust priors
}.
% \lichongyi{the illposeness is not caused by weak-generalized hand-craft priors. Also weak-generalized is not a correct word.}
%
Concretely, 1) collecting large-scale and perfectly aligned hazy images with the clean counterpart is incredibly difficult, if not impossible. 
Thus, most of the existing deep models use synthetic data for training, in which the hazy images are generated using Eq. \eqref{eq:physical_model}, leading to the neglect of multiple degradation factors.
There are some real hazy image datasets~\cite{ancuti2019dense, ancuti2020nh} with paired data, but the size and diversity are insufficient. Moreover, these datasets deviate from the hazy images captured in the wild.
These shortcomings inevitably decrease the capability of deep models in real scenes. 
2) Real image dehazing is a highly ill-posed issue.
Generally, addressing an uncertain mapping problem often needs the support of priors.
However, it is difficult to obtain robust priors that can cover the diverse scenes of real hazy images, which also limits the performance of dehazing algorithms.
Recently, many studies for real image dehazing try to solve these two issues by domain adaptation from the perspective of data generation~\cite{shao2020domain, yang2022self} or priors guidance~\cite{li2019semi, chen2021psd}, but still cannot obtain desirable results.
In this work, we present a new paradigm for real image dehazing motivated by addressing the above two problems.
To obtain large-scale and perfectly aligned paired training data, we rethink the degradation of hazy images by observing amounts of real hazy images and propose a novel data generation pipeline considering multiple degradation factors.
In order to solve the uncertainty of the ill-posed issue, we attempt to train a VQGAN~\cite{esser2021taming} on high-quality images to extract more robust high-quality priors (HQPs).
The VQGAN only learns high-quality image reconstruction, so it naturally contains the robust HQPs that can help hazy features jump to the clean domain. 
The observation in Sec.~\ref{sec:pretrain} further verifies our motivation.
Thus, we propose the \textbf{R}eal \textbf{I}mage \textbf{D}ehazing network via high-quality \textbf{C}odebook \textbf{P}riors (RIDCP). %
The codebook and decoder of VQGAN are fixed to provide HQPs.
Then, RIDCP is equipped with an encoder that helps find the correct HQPs, and a new decoder that utilizes the features from the fixed decoder and produces the final result.
Moreover, we propose a novel Normalized Feature Alignment (NFA) that can mitigate the distortion and balance the features for better fusion.
In comparison to previous methods~\cite{gu2022vqfr, zhou2022codeformer, chen2022real} that introduce codebook for image restoration, we further design a unique real domain adaptation strategy based on the characteristics of VQGAN and the statistical results.
Intuitively, we propose Controllable HQPs Matching (CHM) operation that replaces the nearest-neighbour matching by imposing elaborate-designed weights on the distances between features and HQPs during the inference phase.
The weights are determined by a controllable parameter $\alpha$ and the statistical distribution gap of HQPs activation in Sec.~\ref{sec:matching}. 
By adjusting $\alpha$, the distribution of HQPs activation can be shifted.
Moreover, we present a theoretically feasible solution to obtain the optimal $\alpha$ by minimizing the Kullback-Leibler Divergence of two probability distributions.
More significantly, the value of $\alpha$ can be visually reflected as the enhancement degree as shown in Figure~\ref{fig:intro_compare}(d),
and users are allowed to adjust the dehazing results as per their preference.
Our CHM is effective, flexible, and explainable.
% }
%%%%%%%%%%%%%%%%%%%%%%%%%%%%%%%%%%%%%%%%%%%%%%%%%%%%%%%%%%%%%%%%%%

% \dzp{Different from the application of codebook in other image restoration tasks,
% real image dehazing suffers from the domain gap between high-quality training data and low-quality hazy input,
% resulting in the biased code matching and thus confining the reconstruction ability of this technique.
% In order to manually ameliorate the activation distribution of HQPs, 
% we propose Controllable HQPs Matching (CHM) mechanism that imposes an elaborate-designed weight on the distance between features and HQPs during inference phase.
% Based on the statistical observation in Sec.~\ref{sec:matching}，
% the weight combines a controllable parameter $\alpha$ and the frequency gap of each code activated by clear and hazy features.
% By adjusting $\alpha$,
% the distribution of activated HQPs can be shifted,
% visually reflected as the enhancement degree as shown in Figure~\ref{fig:intro_compare}(d).
% More significantly,
% we present a theoretically feasible solution to obtain the optimal $\alpha$ to minimize the domain gap statistically.
% Perceptual results further demonstrate that our CHM is effective, flexible and explainable.
% }

Compared with the state-of-the-art real image dehazing methods, \eg, DAD~\cite{shao2020domain} and PSD~\cite{chen2021psd}, only the proposed RIDCP can effectively process the hazy images captured in the wild while generating adjustable results, which are shown in Figure~\ref{fig:intro_compare}.
The contributions of our work can be summarized as follows.
\begin{itemize}
    \item We present a new paradigm to push the frontier of deep learning-based image dehazing towards real scenes. 
    \item We are the first to leverage the high-quality codebook prior in the real image dehazing task. The controllable HQPs matching operation is proposed to overcome the gap between synthetic and real domains and produce adjustable results.
    \item We re-formulate the degradation model of real hazy images and propose a phenomenological degradation pipeline to simulate the hazy images captured in the wild.
\end{itemize}

\section{Related Work}
\subsection{Single Image Dehazing}
\noindent\textbf{Image Dehazing.}
The early attempts at single image dehazing
consider estimating the parameters of the atmosphere scattering model presented in Eq.~(\ref{eq:physical_model}) 
by priors on haze-free images~\cite{tan2008visibility, he2010single,zhu2014single,fattal2014dehazing,berman2016non},
% among which the widely used ones are contrast maximization~\cite{tan2008visibility}, dark channel prior~\cite{he2010single}, color attenuation prior~\cite{zhu2014single}, color lines prior~\cite{fattal2014dehazing} and non-local prior~\cite{berman2016non}.
These methods have achieved impressive results.
However, the handcrafted priors based on empirical observations are hard to perform well in diverse scenarios.
% However, the handcraft priors can not always perform well on complex real scenes,
For example, the assumption of DCP~\cite{he2010single} is not available  in the sky region.
The proposed method obtains the priors of high-quality images by pre-training a discrete codebook on large-scale datasets, which is more reliable and comprehensive.
% For instance, in dark channel prior~\cite{he2010single}, He~\etal proposed that the minimum in the RGB channel should be close to zero except for sky region.
% Zhu~\etal 

With the development of deep learning techniques,
% the deep neural networks boost the performance of single image dehazing significantly.
%
how to use data-driven ideology to remove haze gains a lot of attention.
At the early stage, many studies~\cite{cai2016dehazenet, ren2016single, li2017aod} try to adopt convolutional neural networks (CNNs) to estimate the parameters of the degradation model in Eq.~(\ref{eq:physical_model}).
%
% DehazeNet~\cite{cai2016dehazenet} is a pioneer work of data-driven approaches,
% which estimates transmission map for haze removal.
% %
% Ren~\etal~\cite{ren2016single} propose a multi-scale network
% to generate the correspondence transmission map of hazy input.
% %
% Li~\etal~\cite{li2017aod} reformulate the atmospheric scattering model and design a light weight CNNs for parameters estimation.
%
In addition, in order to avoid accumulated errors in parameters estimation,
some end-to-end networks~\cite{liu2019griddehazenet, qin2020ffa, dong2020multi, guo2022image, ye2021perceiving} are proposed to directly estimate the haze-free image.
The above learning-based methods have achieved excellent performance on synthetic datasets. However, their significant performance drop on real-world data urgently needs to be solved.
% Liu~\etal~\cite{liu2019griddehazenet} proposed GridDehazeNet which is composed of pre-processing, backbone, and post-processing.
% %
% In FFANet~\cite{qin2020ffa}, a module combines channel attention and pixel attention is proposed for feature fusion.
% %
% Dong~\etal~\cite{dong2020multi} design a dense feature fusion mechanism based on U-Net architecture.
% %
% Guo~\etal~\cite{guo2022image} combine the features extracted by CNNs and Transformer to boost the performance.
% %
% Ye~\etal~\cite{ye2021perceiving} propose a two-stage architecture to recover hazy input with the guidance of density map.
%

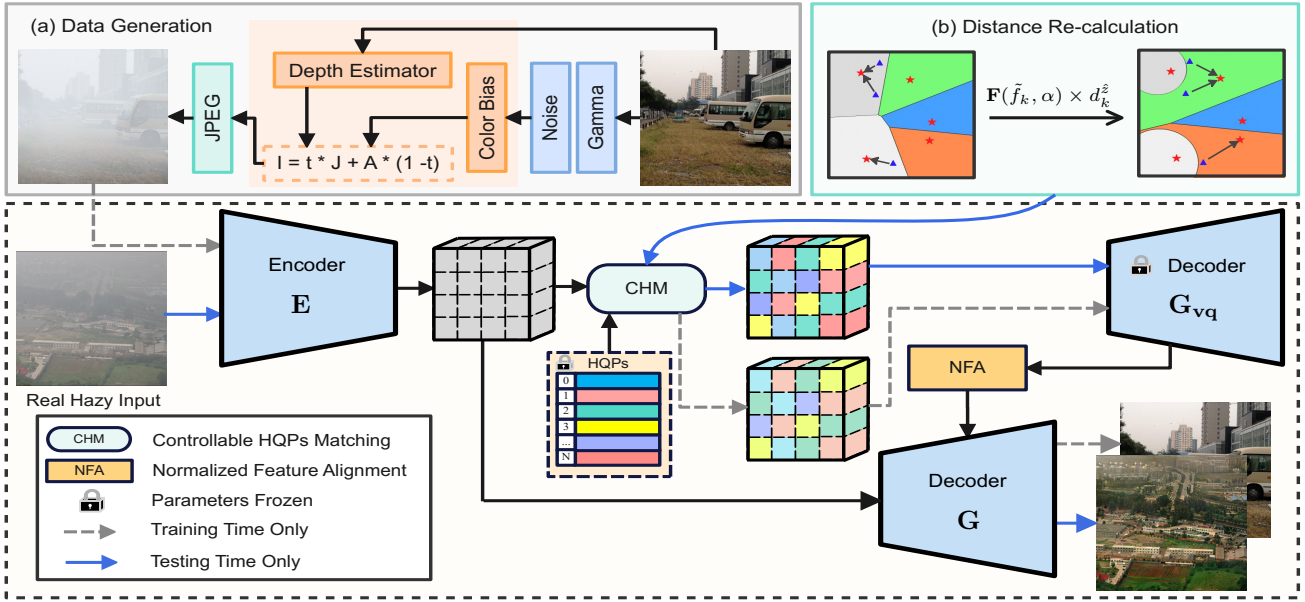
\begin{figure*}[!t]
    \centering
    \includegraphics[width=\textwidth, height=8cm]{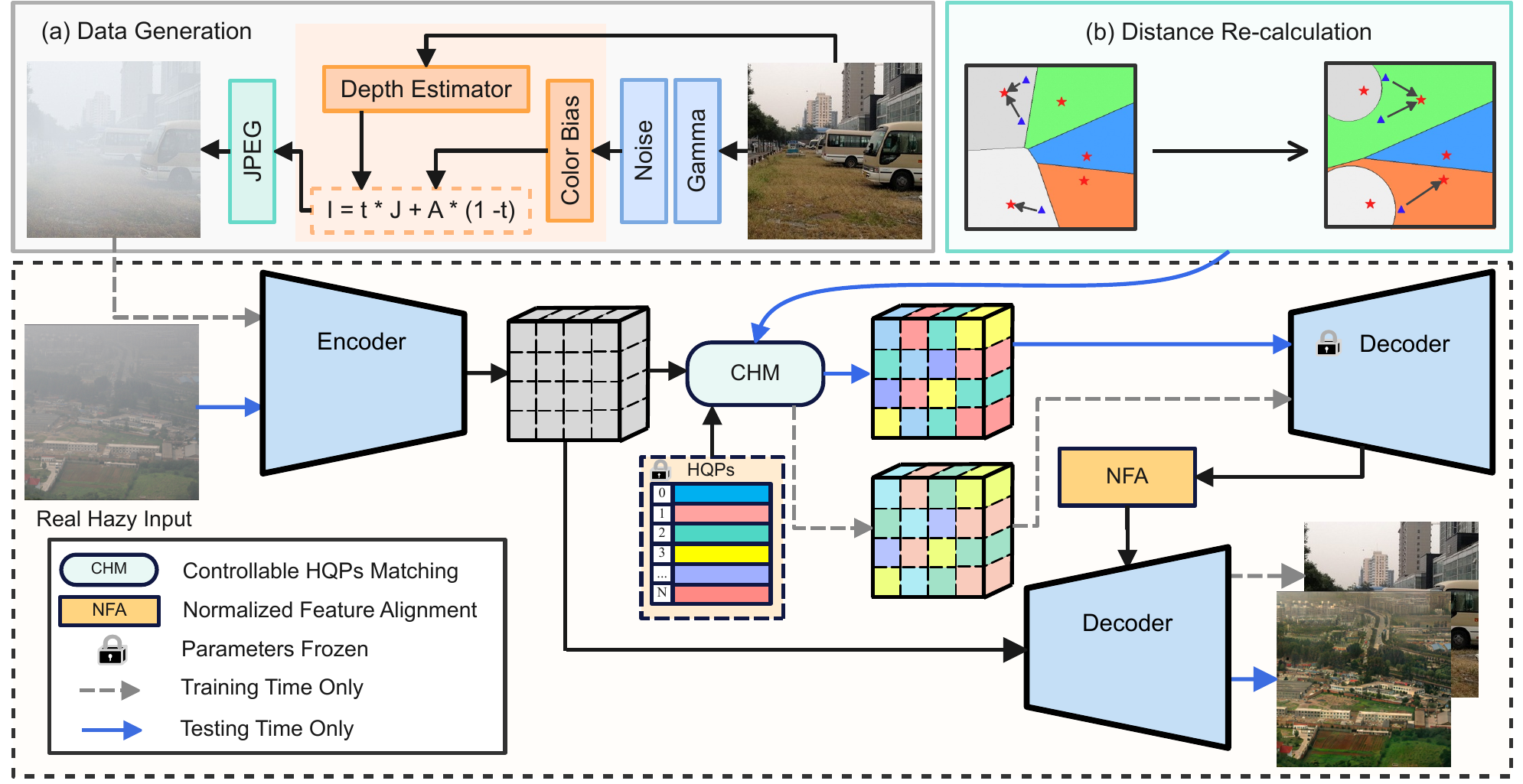}
    \put(-53, 110){{$\mathbf{G_{vq}}$}}
    \put(-133, 28){{$\mathbf{G}$}}
    \put(-385, 110){{$\mathbf{E}$}}
    \put(-122, 190){\scriptsize{$\mathbf{F}(\Tilde{f}_k, \alpha) \times d^{\hat{z}}_{k}$}}
    \vspace{-0.25cm}
    \caption{Overview of our RIDCP. 
    During the training phase, we train the dehazing network on the data synthesized by our data generation pipeline, as illustrated in (a).
    The network is based on the pre-trained HQPs codebook and the corresponding decoder $\mathbf{G}_{vq}$ of VQGAN.
    We also design the Controllable HQPs Matching (CHM) operation for real domain adaptation by re-calculating the distance $d^{\hat{z}}_{k}=||\hat{z}-z_k||$ between features and HQPs.
    (b) represents the distance re-calculation with two Voronoi diagrams, where the colored cells indicate belonging to better HQPs and the gray cells vice versa. Triangles represent features and star points represent HQPs. It can be seen that after the distance recalculation points that originally belonged to the gray cells are forced to be assigned to the colored cells by our CHM.
    }
    \vspace{-0.5cm}
    \label{fig:overview}
\end{figure*}

\noindent\textbf{Real Image Dehazing.}
Recently, some works pay attention to real image dehazing.
One research line is to utilize GANs~\cite{goodfellow2014generative} for generating hazy data that fits the real haze domain.
Shao~\etal~\cite{shao2020domain} design a domain adaptation strategy based on the framework of CycleGAN~\cite{CycleGAN2017}.
Yang~\etal~\cite{yang2022self} propose an unpaired dehazing framework named D4. It can estimate the scene depth of hazy images and generate hazy data with different thicknesses to benefit dehazing model training. 
However, GANs are easy to produce artifacts in the generated results, which is harmful to training models.
Another research line aims to introduce prior knowledge by loss functions or network architectures.
Li~\etal~\cite{li2019semi} propose a semi-supervised pipeline that adopts prior-based loss functions to train networks on the real dataset.
PSD~\cite{chen2021psd} adds a physical-based sub-network on the pre-trained dehazing model and further proposes a prior loss committee to fine-tune the network on real-world data in an unsupervised manner.
Nevertheless, directly using handcrafted priors cannot avoid the inherent flaws of prior-based methods.
In our study, we investigate overcoming the weaknesses of both types of real image dehazing methods by proposing a novel data generation pipeline and exploiting the latent high-quality priors.

\subsection{Discrete Codebook Learning}
Recently, a vector-quantized auto-encoder framework was proposed in VQ-VAE~\cite{van2017neural}, which learns a discrete codebook in latent space. The discrete representation effectively addresses the ``posterior collapse'' issue in auto-encoder~\cite{hinton1993autoencoders} architecture.
VQGAN~\cite{esser2021taming} further improves the perceptual quality of reconstructed results by introducing adversarial supervision for codebook learning.
%
% The learned discrete codebook help boost the performance in many low-level vision tasks including face restoration~\cite{gu2022vqfr, zhou2022codeformer}, super-resolution~\cite{chen2022real} and image colorization~\cite{huang2022unicolor}.
The learned discrete codebook helps boost the performance in many low-level vision tasks including face restoration~\cite{gu2022vqfr, zhou2022codeformer} and image super-resolution~\cite{chen2022real}. Gu~\etal~\cite{gu2022vqfr} introduce the vector quantization technique to face restoration and design a parallel decoder to achieve a balance between visual quality and fidelity. Zhou~\etal~\cite{gu2022vqfr} cast blind face restoration as a code prediction task, and propose a Transformer-based prediction network to replace the nearest-neighbor matching operation for better matching the corresponding code. FeMaSR~\cite{chen2022real} extends the discrete codebook learning to blind super-resolution. 
%
% However, the complex degradation and high degree information loss in hazy data cause problems for correct matching in codebook.
%
% Motivated by this, we give a detail discussion and a effective solution for this issue.
Motivated by the exciting performance of these approaches, we are the first to leverage the high-quality codebook prior for real image dehazing. A novel and controllable HQPs matching operation is proposed to further bridge the gap between our synthetic data and real data, which is inevitable for real scenes.
% \begin{figure}[!t]
%     \centering
%     \includegraphics[width=\linewidth]{figures/data_pipeline.pdf}
%     \vspace{-0.7cm}
%     \caption{Overview of our data generation pipeline.}
%     \label{fig:data_pipeline}
%     \vspace{-0.5cm}
% \end{figure}

\section{Data Preparation for Real Image Dehazing}
Redesigning the pipeline of data generation has been demonstrated as an effective way for solving real-world low-level vision tasks~\cite{wang2021real, zhang2021designing, wang2021rain}.
Based on these works, we consider various degradation factors when synthesizing paired data for training the dehazing network, which can mitigate the domain gap with the real data.
For conciseness, we represent Eq.~(\ref{eq:physical_model}) as $I(x) = \mathcal{P}(J(x), t(x), A)$.
The formation of the hazy image can be written as:
\begin{equation}
    I(x) = JPEG(\mathcal{P}(J(x)^{\gamma} + \mathcal{N}, e^{\beta \hat{d}(x)}, A + \Delta{A})).
    \vspace{-0.1cm}
    \label{eq:data_pipeline}
\end{equation}
The details of Eq.~(\ref{eq:data_pipeline}) are introduced as follows:
\noindent\textbf{Poor light condition.}
$\gamma\in[1.5, 3.0]$ is a brightness adjustment factor and $\mathcal{N}$ is the Gaussian noise distribution.
These two components can simulate poor light conditions that frequently occur in hazy weather.
\noindent\textbf{Transmission map.}
As a key parameter in the degradation model, we adopt the depth estimation algorithm~\cite{he2022ra} to estimate depth map $d(x)$ and use $\beta \in [0.3, 1.5]$ to control the haze density.
\noindent\textbf{Colorful haze.}
To obtain diverse hazy images, the color bias of atmosphere light is considered, which is implemented by a  three-channel vector $\Delta A \in [-0.025, 0.025]$.
The range of $A$ is in the range of $[0.25, 1.0]$.
\noindent\textbf{JPEG compression.}
We observe that dehazing algorithms amplify the JPEG artifacts.
It is desirable to remove such artifacts while dehazing.
$JPEG(\cdot)$ denotes JPEG compression in the final results.
%

% We select $500$ clean images to build the paired data.
% Different from the previous work~\cite{li2019benchmarking}, the hazy image is generated on-the-fly during training phase, which aims to make networks learn more variegation knowledge \lichongyi{cannot understand the words `variegation knowledge'. }.
% Additionally, low light and JPEG compression appear with 50\% probability in the proposed pipeline.
We select $500$ clean images to build the paired data, and the hazy data is generated on-the-fly during the training phase. Additionally, low light and JPEG compression appear with 50\% probability in the proposed pipeline.
\section{Methodology}
The key idea of our work is to adopt a discrete codebook that introduces high-quality priors (HQPs) into the dehazing network.
The overall framework of the proposed method is illustrated in Figure~\ref{fig:overview}.
The training phase can be divided into two stages.
In the first training stage, we pre-train a VQGAN~\cite{esser2021taming} on high-quality data, obtaining a latent discrete codebook $\mathcal{Z}$ with HQPs and the correspondence decoder $\mathbf{G}_{vq}$ (Sec.~\ref{sec:pretrain}).
In the second stage, our RIDCP based on the pre-trained VQGAN is trained on hazy images generated by the proposed synthesis pipeline (Sec.~\ref{sec:dehazing}).
Moreover, in order to help the network find more accurate code, we propose a controllable adjustment feature matching strategy based on code activation distribution on high-quality images (Sec.~\ref{sec:matching}).
Besides, the details of training objectives can be found in supplementary materials.

\subsection{Latent Codebook for High-quality Priors}
\label{sec:pretrain}

We first introduce how VQGAN works briefly.
Given a high-quality image patch $x$,
which is the input of the VQGAN encoder $\mathbf{E}_{vq}$  
and the corresponding outputs are the latent features $\hat{z}$.
Then each ``pixel'' $\hat{z}_{ij}$ of $\hat{z}$ will be matched to the nearest HQPs in codebook $\mathcal{Z} \in \mathbb{R}^{K \times n}$ and then obtain the discrete representation $z^q_ {ij}$, which can be written as:
\begin{equation}
\vspace{-0.1cm}
    z^q_{ij} = \mathcal{M}(\hat{z}_{ij}) = arg\min\limits_{z_k\in \mathcal{Z}}(||\hat{z}_{ij} - z_{k}||_2),
    \vspace{-0.1cm}
    \label{eq:matching}
\end{equation}
where $K$ denotes the codebook size, $n$ is the channel number of $\hat{z}$, and $\mathcal{M}(\cdot)$ represents the matching operation.
Finally, the input $x$ is reconstructed by $\mathbf{G}_{vq}$:
\begin{equation}
    \vspace{-0.1cm}
    x^\prime = \mathbf{G}_{vq}(z^q) =  \mathbf{G}_{vq}(\mathcal{M}(\mathbf{E}_{vq}(x))),
    \vspace{-0.1cm}
\end{equation}
where $x^\prime$ is the reconstructed result.

% \textbf{Training Objectives.}
% \wrq{I suggest putting this section in supplementary materials.}
% Since the operation in Eq.~(\ref{eq:matching}) is non-differentiable,
% VQGAN is end-to-end trained by copying the gradients of $\mathbf{G}_{vq}$ to $\mathbf{E}_{vq}$~\cite{bengio2013estimating}.
% %
% The optimization strategy can be divided into two parts, which minimize the loss after reconstruction and after feature matching respectively.
% For first part, the loss function can be formulated as:
% \begin{equation}
%     \mathcal{L}_{rec} = ||x^\prime - x||_1 + \mathcal{L}_{per} + \mathcal{L}_{adv},
% \end{equation}
% where $\mathcal{L}_{per}$ and $\mathcal{L}_{adv}$ are perceptual loss~\cite{ledig2017photo} and adversarial loss~\cite{ledig2017photo}, respectively. And for codebook optimization, the loss function can be written as:
% \begin{equation}
% \begin{split}
%     \mathcal{L}_{codebook} = ||sg(\hat{z}) - z^q||^2_2 + \beta ||sg(z^q) - \hat{z}||_2^2 \\+ \gamma||CONV(z^q) - \phi(x)||_2^2,
% \end{split}
% \end{equation}
% where $sg(\cdot)$ is the stop-gradient operation, and $\beta=0.25, \gamma=0.1$ respectively. 
% The last term of $\mathcal{L}_{codebook}$ is a semantic guided regularization term follow~\cite{chen2022real}, where $CONV$ is a simple convolutional layer and $\phi$ is the pretrained VGG19~\cite{journals/corr/SimonyanZ14a}. Finally, the total loss of VQGAN is:
% \begin{equation}
%     \mathcal{L}_{vq} = \mathcal{L}_{rec} + \mathcal{L}_{codebook}.
% \end{equation}

\begin{figure}[t]
    \centering
    \begin{subfigure}{.48\linewidth}
        \includegraphics[width=\linewidth, height=2.3cm]{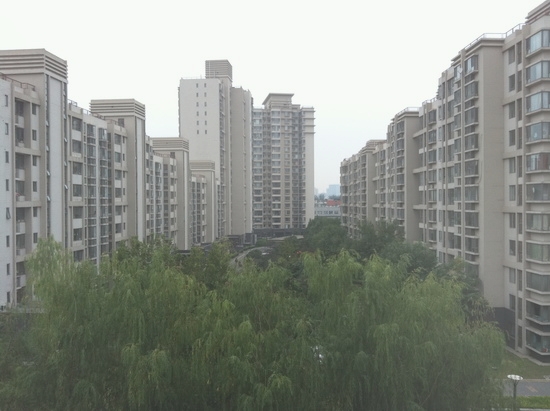}
        \put(-85, -10){\small{(a)~Hazy input}}
        \vspace{-0.25cm}
    \end{subfigure}
    \begin{subfigure}{.48\linewidth}
        \includegraphics[width=\linewidth, height=2.3cm]{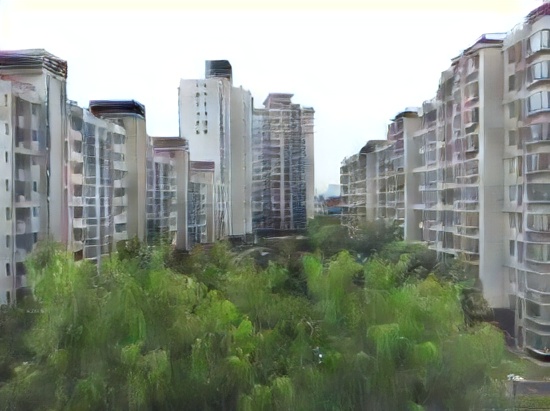}
        \put(-90, -10){\small{(b)~Reconstruction}}
        \vspace{-0.25cm}
    \end{subfigure}
    \caption{Result reconstructed by the pre-trained VQGAN. The haze is removed but distorted textures are introduced.}
    \label{fig:ob1}
    \vspace{-0.55cm}
\end{figure}

\noindent\textbf{Observation 1.}
To understand the potential of the HQPs in the codebook and better utilize it, we made some observations on the results reconstructed by the pre-trained VQGAN.
As illustrated in Figure~\ref{fig:ob1},
our VQGAN can remove the thin haze and recover vivid color for the hazy input without fine-tuning.
We analyze that using HQPs in a matching manner can replace the degraded feature so help it jump to the high-quality domain.
% \lichongyi{reduce? jump is not correct. } the negative effects of haze. 
%
% However, the dehazing ability of VQGAN is limited due to the difficulty of matching the correct code, and some distorted textures are produced since the information loss during vector-quantized phase.
% %
% Therefore, it is natural to think that our next step is to train an encoder $\mathbf{E}$ that can help priors matching, and a decoder $\mathbf{G}$ that can eliminate the artifacts while using the information from HQPs.
%
However, the dehazing ability of VQGAN is limited due to the difficulty in matching the correct code.
Moreover, some distorted textures are produced because of the information loss during the vector-quantized phase.
Thus, directly adopting the features from $\mathbf{G}_{vq}$ is suboptimal. 
It is intuitive that our next step is to train an encoder $\mathbf{E}$ that can help priors matching, and a decoder $\mathbf{G}$ that can utilize the features reconstructed from HQPs.

\subsection{Image Dehazing via Feature Matching}
\label{sec:dehazing}
Based on the observation in Sec.~\ref{sec:pretrain}, image dehazing is decoupled into two sub-tasks:
matching correct code and removing texture distortion.

\noindent\textbf{Encoder for HQPs Matching.}
We follow SwinIR~\cite{liang2021swinir} which shows its powerful feature extraction ability for image restoration to design our encoder $\mathbf{E}$.
Specifically, the shallow feature extraction head consists of a stack of residual layers~\cite{he2016deep} and  $4\times$ downsamples the features. 
Then $4$ residual swin transformer blocks~\cite{liu2021swin}
are followed, which serve as the deep feature extraction module.

\noindent\textbf{Decoder with Normalized Feature Alignment.}
We propose the Normalized Feature Alignment (NFA) to help the decoder utilize the features reconstructed from HQPs.
Firstly, VQGAN tends to decrease results' fidelity  due to the information loss brought by vector-quantized operation~\cite{gu2022vqfr, zhou2022codeformer}.
%, which has been demonstrated in previous works.
%
Our solution is to eliminate the distortion by the guidance of features before HQPs matching.
Specifically, in $i$th layer, we adopt the deformable convolution~\cite{dai2017deformable} to align the features $F^i_{vq}$ from $\mathbf{G}_{vq}$ with the features $F^i$ from $\mathbf{G}$, which can be written as:.
\begin{equation}
    F^i_{w} = DCONV(F^i_{vq}, CONV(Concat(F^i_{vq}, F^i))),
    \vspace{-0.1cm}
\end{equation}
where $F^i_{w}$ denotes the features after warping and $DCONV$ is the deformable convolutional layer.
$CONV$ is the convolutional layer for offset generation.
In addition, we notice that the ratio of the values of $F^i_{w}$ and $F^i$ is not stable, resulting in an inadequate combination.
Thus, we balance the contributions of each by forcing them to be in the same order of magnitude, which can be written as:
\begin{equation}
    F^i = F^i + \frac{\sum F^i}{\sum F^i_w} F^i_w. 
\end{equation}
% The impact of the proposed NFA is detailed in Sec.~\ref{sec:ablation}.

\begin{figure}[t]
    \centering
    \includegraphics[width=\linewidth]{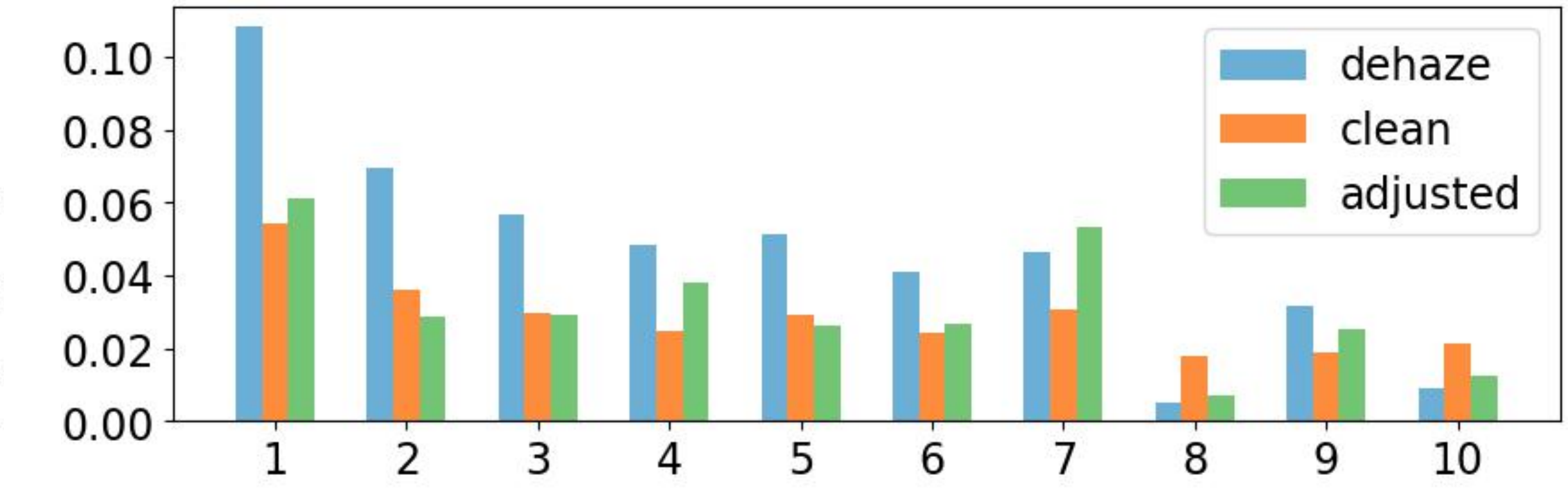}
    \put(-240, 5){\footnotesize{\rotatebox{90}{Activation Frequency}}}
    \put(-150, -5){\footnotesize{Top ten $|f_h - f_c|$ codes}}
    \vspace{-0.2cm}
    \caption{
        Code activation frequencies under different situations. 
        `dehaze' denotes inputting real hazy images to dehazing network
        and `clean' denotes feeding clean images into the pre-trained VQGAN network.
        `adjusted' denotes our RIDCP equipped with the CHM under the recommended parameter with real hazy inputs. 
    }
    \label{fig:ob2}
    \vspace{-0.55cm}
\end{figure}

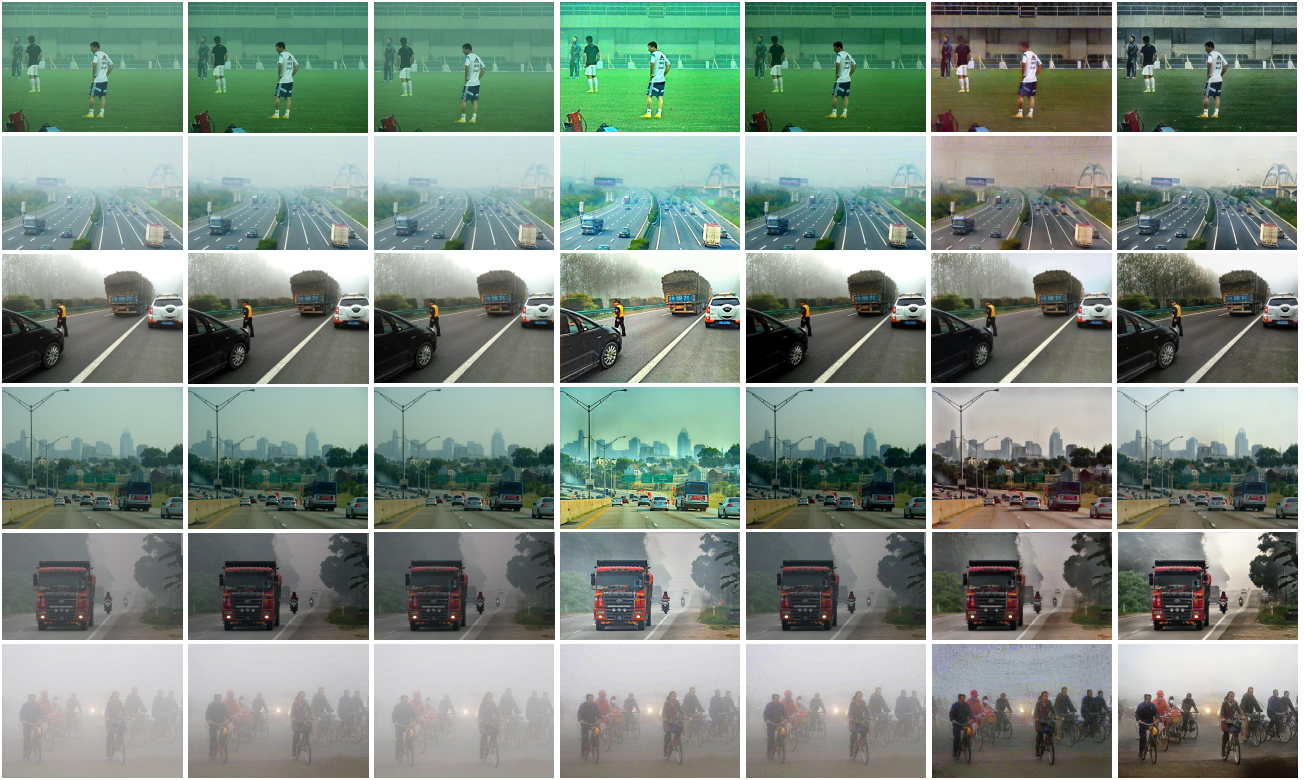
\begin{figure*}[t]
    \centering
    \includegraphics[width=\linewidth]{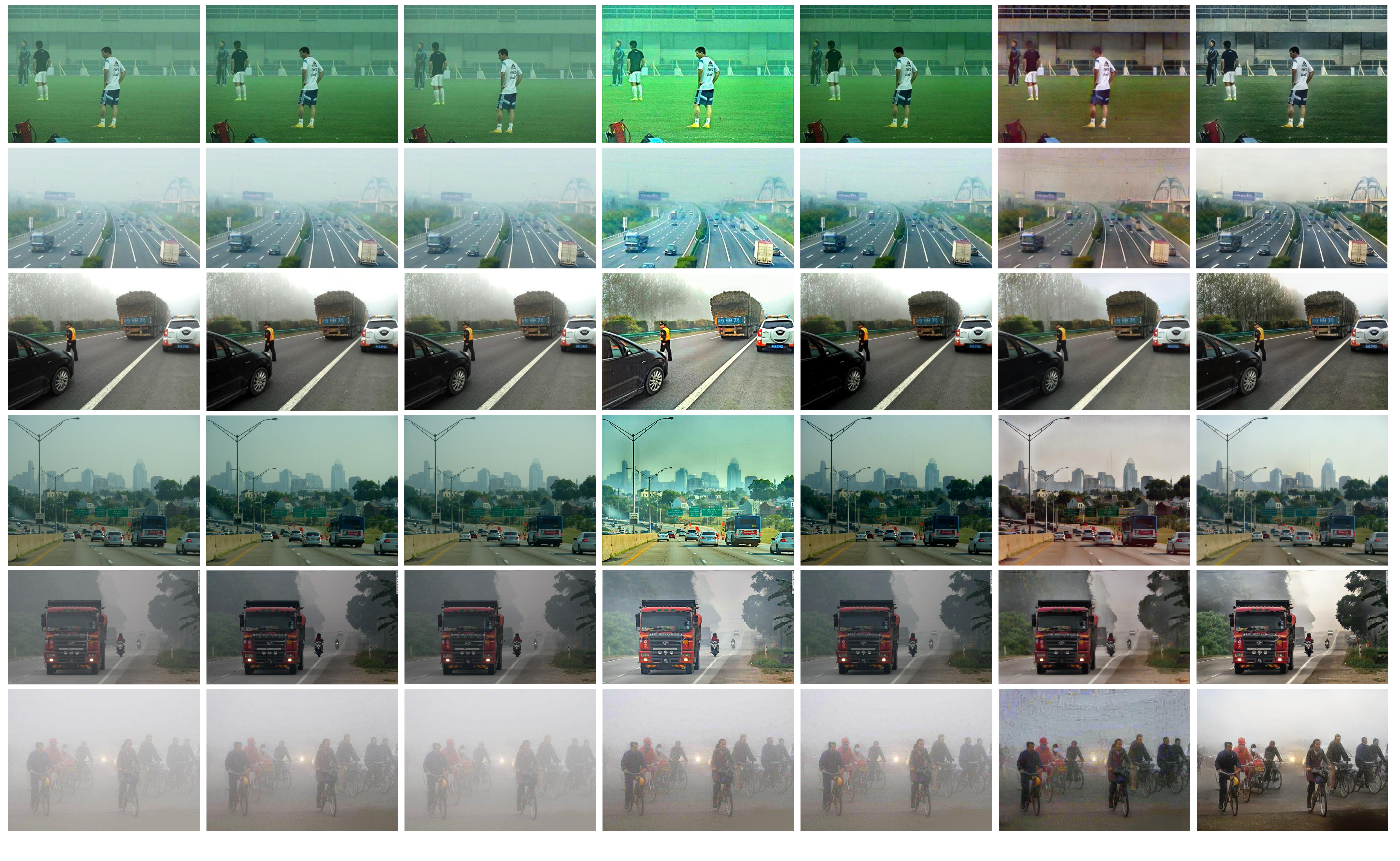}
    \put(-487.5, 0){\small{(a)~Hazy image}}
    \put(-421.5, 0){\small{(b)~MSBDN~\cite{dong2020multi}}}
    \put(-351, 0){\small{(c)~Dehamer~\cite{guo2022image}}}
    \put(-270, 0){\small{(d)~PSD~\cite{chen2021psd}}}
    \put(-199, 0){\small{(e)~D4~\cite{yang2022self}}}
    \put(-133, 0){\small{(f)~DAD~\cite{shao2020domain}}}
    \put(-59, 0){\small{(g)~RIDCP}}
    \vspace{-0.2cm}
    \caption{Visual comparison on RTTS dataset~\cite{li2019benchmarking}.}
    \label{fig:RTTS_comparison}
    \vspace{-0.4cm}
\end{figure*}

\subsection{Controllable HQPs Matching Operation}
\label{sec:matching}
\noindent\textbf{Observation 2.}
Our RIDCP achieves relatively satisfactory results with the help of $\mathbf{E}$ and $\mathbf{G}$.
However, there are still limitations, \eg, low color saturation in some challenging real data. 
Rather than the HQPs that already show a strong capability in reconstructing vivid results (see Observation 1), the main reason is the difficulty in finding correct HQPs, which is caused by the domain gap between synthetic data and real data.
Although the domain gap is drastically reduced by our synthesis pipeline than previous works~\cite{li2019benchmarking, zheng2021ultra}, it is still impossible to cover all real-world hazy conditions by our pipeline. 

% unavoidable domain gap makes it troublesome to find correct HQPs, 

% \lele{It is difficult to find the correct HQPs through L1/L2 loss, which is an indirect constraint.}
%
To verify our claim, we made an observation as follows.
We randomly collect 200 high-quality clean images as input of the pre-trained VQGAN and compute the activation frequency $f_{c} \in \mathbb{R}^K$ of each code.
Similarly, 200 real hazy images are fed to the dehazing network to compute the frequency $f_{h} \in \mathbb{R}^K$.
Figure~\ref{fig:ob2} illustrates the activation frequencies of the codes with the top ten largest differences between $f_{h}$ and $f_{c}$.
We can see a significant distribution shift.
The observation proves that the unavoidable domain gap results in a divergent matching for HQPs.
Thus, HQPs still have unexplored potential.
% which is also the next step we need to exploit.
% even though the proposed data generation pipeline can simulate real data better than previous work~\cite{?},
% the dehazing network still needs improvement in adapting to real data domain,
% which is also the next step we need to address.
%
% Thus, we make some efforts to help the network adapt to real domain based on frequency difference.

\noindent\textbf{Controllable Matching via Distance Re-calculation.}
Based on the above observation, 
it is indispensable to match better HQPs when encountering real hazy images, 
\ie, priors with high frequency on clear images.
%
% which can make the activation probability of HQPs on real hazy images $p_{h}$ as close in distribution as possible to that on the clear image $p_{c}$.
%
Two components can affect the HQPs matching, which are the encoder $\mathbf{E}$ and the matching operation $\mathcal{M}(\cdot)$.
Since it is difficult to retrain $\mathbf{E}$ on real hazy images without reference images, 
defining a new matching operation $\mathcal{M}^\prime(\cdot)$ sounds like a reasonable solution.
We propose Controllable HQPs Matching (CHM) that re-calculates distances by assigning different weights during matching phase. 
The CHM can be written as:
\begin{equation}
    \mathcal{M}^\prime(\hat{z})
    = arg\min\limits_{z_k\in Z}(\mathbf{F}(\Tilde{f}_k, \alpha) \times ||\hat{z}-z_k||),
\end{equation}
where $\mathbf{F}(\Tilde{f}_k, \alpha)$ is the function to generate weights based on the frequency difference $\Tilde{f}_k = f_h^k - f_c^k$ and adjusted by a parameter $\alpha$.
There are three objectives in the design of $\mathbf{F}$:
1) Since higher $\Tilde{f}_k$ means less activation is needed, $\mathbf{F}$ should be monotonic with $\Tilde{f}_k$ thus ensuring consistent trend adjustment.
2) $\mathbf{F}(0, \alpha) \equiv 1$ so that HQPs with the same frequencies on clear and hazy data are not adjusted. 
3) The degree of adjustment can be controlled monotonically by $\alpha$, \eg, 
$\forall \Tilde{f}_1 > \Tilde{f}_2, 
\forall \alpha_1 > \alpha_2 \rightarrow
\frac{\mathbf{F}(\Tilde{f}_1, \alpha_1)}{\mathbf{F}(\Tilde{f}_2, \alpha_1)}
>
\frac{\mathbf{F}(\Tilde{f}_1, \alpha_2)}{\mathbf{F}(\Tilde{f}_2, \alpha_2)}$.
Coincidentally, the exponential function has these properties, thus $\mathbf{F}$ can be formulated as;
\begin{equation}
    \mathbf{F}(\Tilde{f}_k, \alpha) = e^{a\times\Tilde{f}_k}.
    \label{eq:control}
\end{equation}
Figure~\ref{fig:overview}(b) adopts two Voronoi diagrams to simulate the changes occurring in the high-dimensional space during feature matching.
As we can see, the points originally belonging to gray cells are matched to the colored cells after distance re-calculation, \ie, finding better HQPs.  

\noindent\textbf{Possible Solution of the Recommended $\boldsymbol{\alpha}$.} 
Our method is able to control the HQPs matching based on the above strategy.
% And the final goal is to find a suitable $\alpha$ to make the probability distribution of the HQPs activation on the real hazy images consistent with the distribution on the clean images.
% According to the law of large numbers, The frequencies $f_c^k, f_h^k$ can be substituted for the corresponding probabilities $P_c(x = z_k), P_h(x = z_k|\alpha)$.
% We calculate the optimal parameter $\hat{\alpha}$ by minimizing the forward Kullback-Leibler Divergence of $P_c(x = z_k)$ and $P_h(x = z_k|\alpha)$, which is also the maximum likelihood estimation of $\alpha$:
The final goal is to find a suitable $\alpha$ to adapt the network to real domain.
According to the law of large numbers, the frequencies $f_c^k, f_h^k$ can be substituted for the corresponding probabilities $P_c(x = z_k), P_h(x = z_k|\alpha)$.
The gap between the dehazing results and the clean domain can be represented by the difference between the two probability distributions.
Thus, the real domain adaptation problem is transferred into calculating an optimal parameter $\hat{\alpha}$ that can minimize the forward Kullback-Leibler Divergence of $P_c(x = z_k)$ and $P_h(x = z_k|\alpha)$, which is also the maximum likelihood estimation of $\alpha$:
\begin{equation}
\begin{aligned}
    \hat{\alpha} 
    &= arg \min \limits_\alpha KL(P_c||P_h) \\ 
    &= arg \min \limits_\alpha \sum \limits_{i=1}^K P_c(x = z_i) \log \frac{P_c(x = z_i)}{P_h(x = z_i|\alpha)}\\
    &= arg \max \limits_\alpha \sum \limits_{i=1}^K P_c(x = z_i) \log P_h(x = z_i|\alpha)\\
    &= arg \max \limits_\alpha \prod \limits_{i=1}^K P_c(x = z_i) P_h(x = z_i|\alpha).
\end{aligned}
\end{equation}
We use a binary search algorithm to iteratively find the approximate optimal solution for $\hat{\alpha}$.
The final determination is $\hat{\alpha}=21.25$ and higher precision calculations have little effect on the results.
% The value of Kullback-Leibler Divergence with different $\alpha$ can be seen in Figure~\ref{fig:find_alpha}.
\textbf{Note that, $\hat{\alpha}$ may not be the determined choice for all cases. One can flexibly adjust $\alpha$ according to their preference.}

% \begin{figure}[t]
%     \flushright
%     \includegraphics[width=\linewidth]{figures/KL.pdf}
%     \put(-120, -5){$\alpha$}
%     \put(-232, 60){\rotatebox{90}{\ttfamily{KL}}}
%     \caption{Caption. \wrq{Need beautify.}}
%     \vspace{-0.3cm}
%     \label{fig:KL_alpha}
% \end{figure}

\vspace{-0.2cm}
\section{Experiments}
\subsection{Datasets}
\noindent\textbf{High-quality Datasets.}
In order to obtain high-quality results from the pre-trained HQPs,
the VQGAN needs to be trained on large-scale datasets
containing high-resolution and texture-sharp images.
In our work, we use DIV2K~\cite{agustsson2017ntire} and Flickr2K~\cite{lim2017enhanced} (containing $4,250$ images) to train the first stage.
Both datasets are widely used in high-quality reconstruction tasks~\cite{liang2021swinir, chen2022real, li2019feedback}.

\noindent\textbf{Real Haze Datasets.}
We qualitatively and quantitatively evaluate our dehazing network on the RTTS dataset~\cite{li2019benchmarking},
which contains over $4,000$ real hazy images with diverse scenes, resolutions, and degradation issues.
Besides, we use Fattal's dataset~\cite{fattal2014dehazing} that includes 31 classic real hazy cases for further visual comparison.

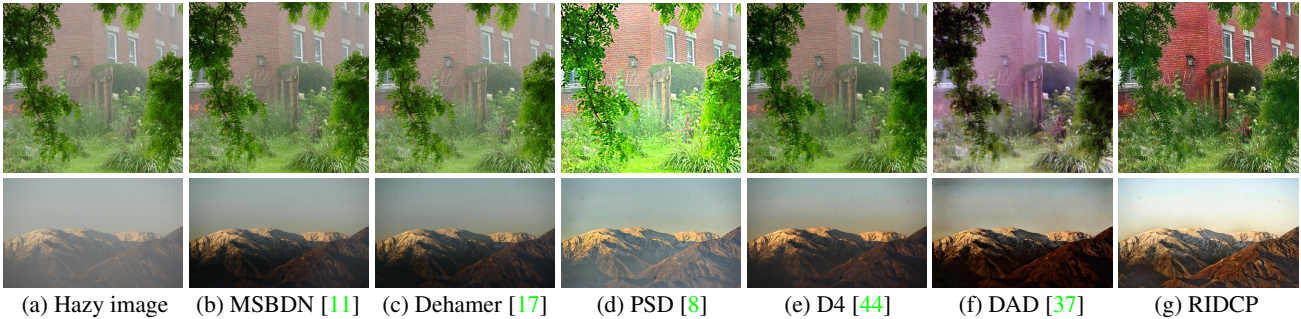
\begin{figure*}[t]
    \centering
    \includegraphics[width=\linewidth]{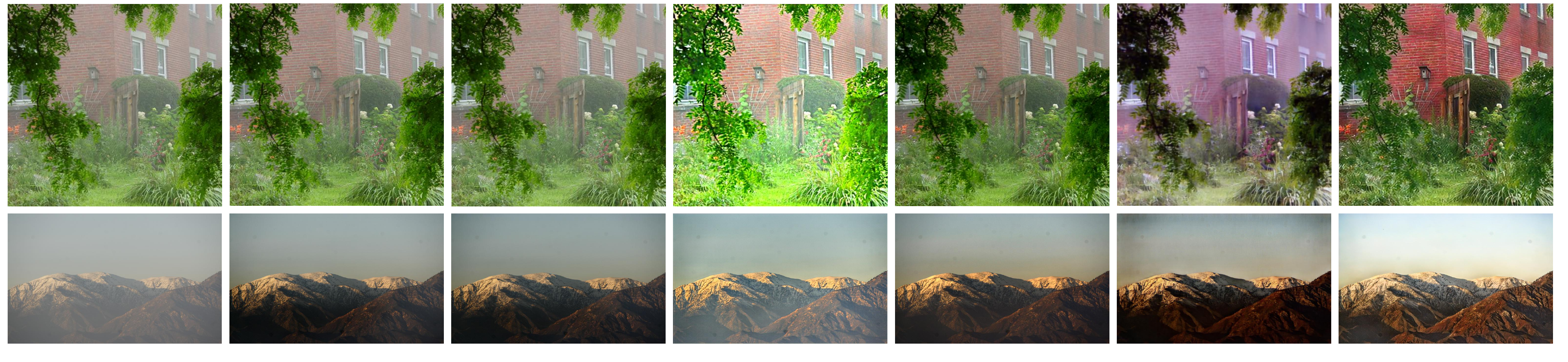}
    \put(-487.5, -3.5){\small{(a)~Hazy image}}
    \put(-421.5, -3.5){\small{(b)~MSBDN~\cite{dong2020multi}}}
    \put(-351, -3.5){\small{(c)~Dehamer~\cite{guo2022image}}}
    \put(-270, -3.5){\small{(d)~PSD~\cite{chen2021psd}}}
    \put(-199, -3.5){\small{(e)~D4~\cite{yang2022self}}}
    \put(-133, -3.5){\small{(f)~DAD~\cite{shao2020domain}}}
    \put(-59, -3.5){\small{(g)~RIDCP}}
    \vspace{-0.15cm}
    \caption{Visual comparison on Fattal's data~\cite{fattal2014dehazing}.}
    \vspace{-0.4cm}
    \label{fig:fattal_comparison}
\end{figure*}

\vspace{-0.2cm}
\subsection{Implementation Details}
For both VQGAN and RIDCP training, we use Adam optimizer with default parameters ($\beta_1=0.9, \beta_2 = 0.99$).
The learning rate is fixed to $0.0001$ during the training phase and the batch size is set to $16$.
For data augmentation, we randomly resize and crop the input into a size of $256\times 256$, and flip it with a half probability.
During the first training stage, our VQGAN is pre-trained on DIV2K and Flickr2K for 350K iterations.
Then, the proposed RIDCP is trained on the data generated by the proposed synthesis pipeline for 10K iterations.
All experiments are implemented with PyTorch framework on 4 NVIDIA V100 GPUs.
The code implemented by MindSpore framework is also provided.

\subsection{Comparison with State-of-the-Art Methods}
We compare the performance of the proposed method with several state-of-the-art dehazing approaches.
The experiments are designed from both quantitative and qualitative perspectives.
Moreover, we also conduct a user study to verify the subjective performance of our method.

\begin{table}[t]
    \centering
    \caption{Quantitative comparison and user study on RTTS dataset. {\color{red}{Red}} indicates the best and {\color{blue}{blue}} indicates the second best. `US' shows the percentage of votes in the user study.}
    \vspace{-0.15cm}
    \renewcommand\arraystretch{0.9}
    \begin{tabular}{c|cccc}
    \toprule
         Method & FADE$\downarrow$ & BRISQUE$\downarrow$ & NIMA$\uparrow$ & US$\uparrow$ \\ \hline
         Hazy image & 2.484 & 37.011 & 4.3250 & 0.030\\
         MSBDN~\cite{dong2020multi} & 1.363 & 28.743 & 4.1401 & 0.046\\
         Dehamer~\cite{guo2022image} & 1.895 & 33.866 & 3.8663 & 0.041 \\
         DAD~\cite{shao2020domain} & 1.130 & 32.727 & 4.0055 & \color{blue}{0.143}\\
         PSD~\cite{chen2021psd} & \color{red}{0.920} & \color{blue}{25.23}9 & \color{blue}{4.3459} & 0.105\\
         D4~\cite{yang2022self} & 1.358 & 33.206 & 3.7239 & 0.079 \\ 
         RIDCP & \color{blue}{0.944} & \color{red}{18.782} & \color{red}{4.4267} & \color{red}{0.556}\\
         \bottomrule
    \end{tabular}
    \vspace{-0.55cm}
    \label{tab:rtts_comparison}
\end{table}

\noindent\textbf{Quantitative Comparison.}
Since there is no ground-truth image in real hazy datasets,
we use some non-reference metrics for quantitative comparison.
We first adopt the Fog Aware Density Evaluator (FADE)~\cite{choi2015referenceless} for haze density estimation.
In addition, two widely-used image quality assessment metrics, BRISQUE~\cite{mittal2011blind} and NIMA~\cite{talebi2018nima} are also included.
The quantitative comparison is conducted on RTTS dataset with two dehazing methods (MSBDN~\cite{dong2020multi} and Dehamer~\cite{guo2022image}) that achieve outstanding performance on synthetic hazy image datasets~\cite{li2019benchmarking},
and three real dehazing approaches (DAD~\cite{shao2020domain}, PSD~\cite{chen2021psd}, D4~\cite{yang2022self}).
The results are illustrated in Table~\ref{tab:rtts_comparison}.
The proposed RIDCP achieves the best in terms of BRISQUE and NIMA,
which gains $25.57\%$ and $1.86\%$ improvements, respectively.
For FADE, our method is ranked second slightly below the PSD.
However, as shown in Figure~\ref{fig:RTTS_comparison}, PSD tends to produce over-enhanced results, which leads to an inaccurate evaluation.
Overall, RIDCP achieves the best results on quantitative metrics, and subsequent experiments will further prove our superiority.

\noindent\textbf{Qualitative Comparison.}
We perform  the qualitative comparison on RTTS and Fattal's datasets,
which is shown in Figure~\ref{fig:RTTS_comparison} and Figure~\ref{fig:fattal_comparison}.
We can observe that Dehamer, MSBDN, and D4 cannot process the real hazy images well.
PSD can produce bright results but the dehazing ability is limited.
DAD is effective in haze removal, while its results suffer from color bias and dark tone.
Our method generates the best perceptual results in terms of brightness, colorfulness, and haze residue compared to other methods.
More results can be found in supplementary materials.

\begin{figure*}[!t]
    \centering
    \includegraphics[width=\linewidth]{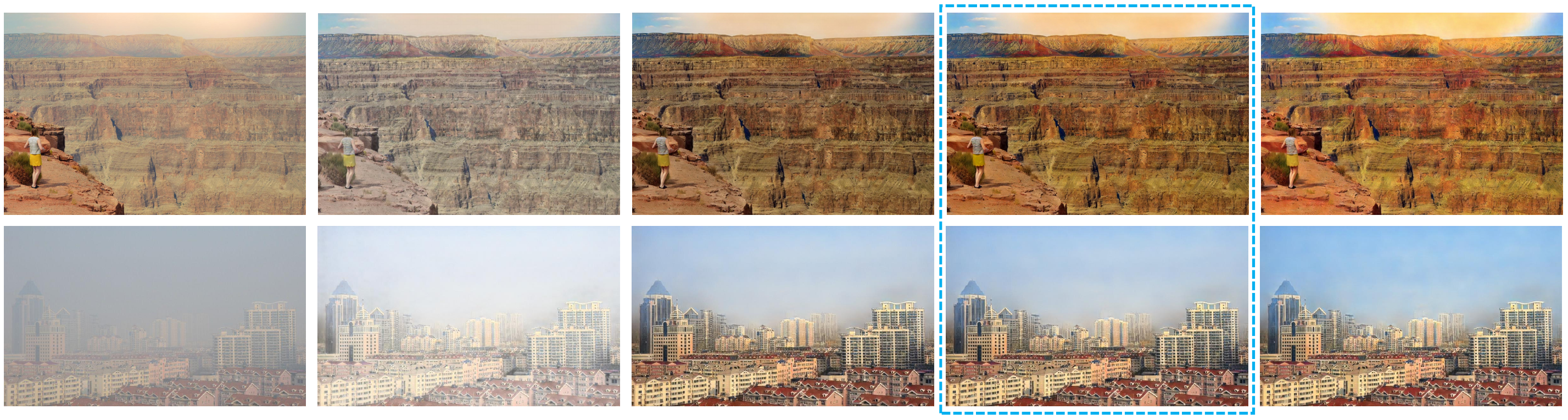}
    \put(-478, -6.5){\small{(a)~Hazy image}}
    \put(-376, -6.5){\small{(b)~$\alpha=-60.0$}}
    \put(-272, -6.5){\small{(c)~$\alpha=0.0$}}
    \put(-178, -6.5){\small{(d)~$\alpha=21.25$}}
    \put(-76, -6.5){\small{(e)~$\alpha=60.0$}}
    \vspace{-0.25cm}
    \caption{Results under different adjustment degrees.
    The proposed CHM allows users to adjust the degree of enhancement from low ($\alpha=-60.0$) to high ($\alpha=60.0$).
    The recommended value ($\alpha=21.25$) facilitates the network to generate the most natural results (surrounded by the blue box). Zoom in for the best view.
    }
    \vspace{-0.45cm}
    \label{fig:weight_adjust}
\end{figure*}

\begin{figure}[t]
    \centering
    \includegraphics[width=\linewidth]{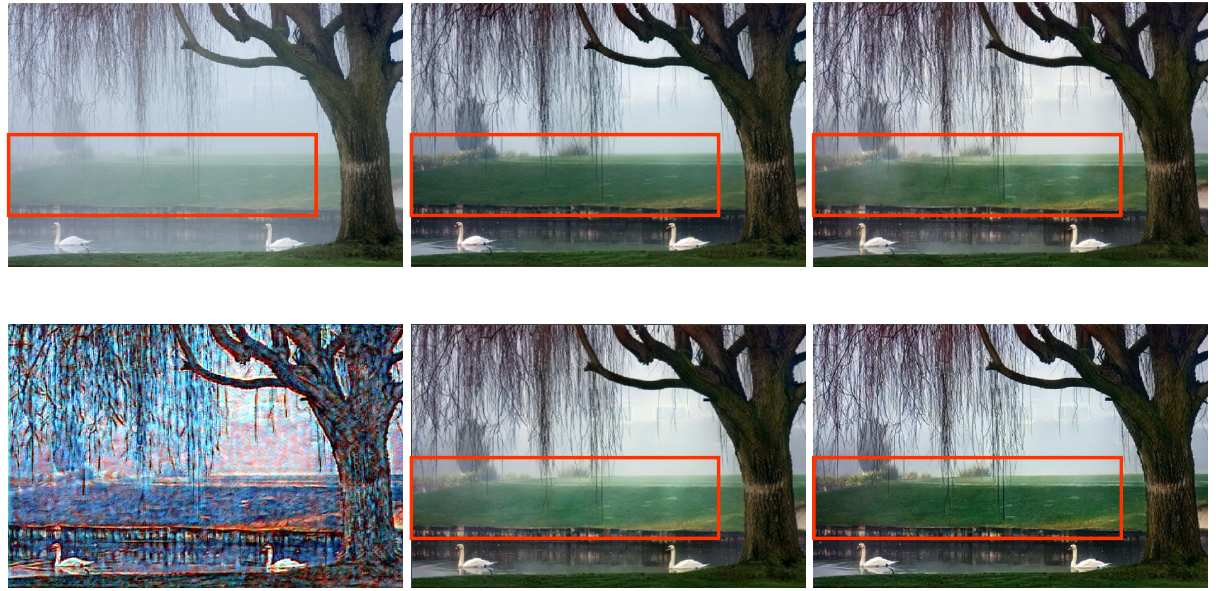}
    \put(-214, 55){\small{(a)~Haze}}
    \put(-147, 55){\small{(b)~w/o fusion}}
    \put(-62, 55){\small{(c)~Addition}}
    \put(-230, -7){\small{(d)~w/o warping}}
    \put(-155, -7){\small{(e)~w/o normalizing}}
    \put(-65, -7){\small{(f)~Full NFA}}
    \vspace{-0.15cm}
    \caption{Ablation results of the proposed NFA.}
    \vspace{-0.65cm}
    \label{fig:nfa_ablation}
\end{figure}
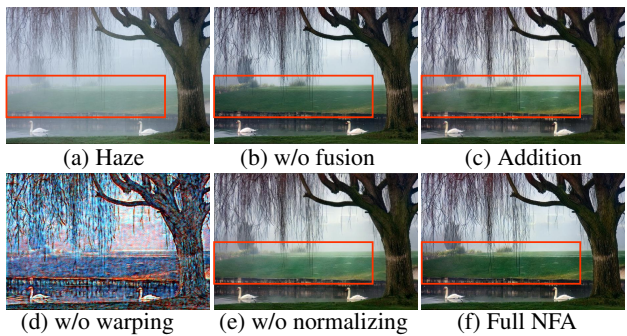

\noindent\textbf{User Study.}
We conduct a user study to evaluate the proposed method subjectively against other methods.
We randomly select 100 images from RTTS dataset for comparison and invite $5$ experts with image processing background and $5$ naive observers as volunteers.
Before the user study, we give the observer three tips:
1) The primary concern is whether the haze is removed, especially the dense haze in the distance.
2) Pay attention to whether the natural color is recovered.
% and over-enhancement does not mean high quality.
%
3) A good method should generate artifact-free results.
Afterward, the images are displayed to the observer group by group. Each group contains the input image and the results generated by different methods.
The observer is required to choose the best one after at least 10 seconds of observation.
We statistic the percentage of each method selected as the best
and the final scores are listed in Table~\ref{tab:rtts_comparison}.
The proposed RIDCP achieves the highest score and is well ahead of the second place, further demonstrating our method's superior dehazing ability.

\subsection{Ablation Study}
\label{sec:ablation}
In order to verify the effectiveness of each key component, we conduct a series of ablation experiments.
Generally, we discuss the effectiveness of CHM, NFA, and the phenomenological degradation pipeline in this section.

\noindent\textbf{Influence of Adjustment Parameter.}
The degree of real domain adaptation is controlled by parameter $\alpha$ in Eq.~\eqref{eq:control}
and one can adjust the final result flexibly by adjusting $\alpha$.
Thus, we are curious about what influence the different $\alpha$ will have.
As Figure~\ref{fig:weight_adjust} shows, the value of $\alpha$ and the image enhancement effect belong to a linear relationship.
More visual-pleasing even over-enhanced results can be produced when $\alpha > 0$.
Interestingly, we can obtain under-enhanced results if $\alpha$ is adjusted in the opposite direction.

\noindent\textbf{Effectiveness of NFA.}
Our NFA can help remove the distorted textures caused by feature matching while preserving the useful information reconstructed from HQPs.
The NFA can be divided into two key parts: warping operation based on deformable convolution and normalize-based addition.
To analyze the role of each part, we propose $4$ variants to replace NFA, which are:
\textit{1) Without any fusion operation.
2) Adding directly.
3) Normalized addition without warping.
4) Warping and direct addition.}
Figure~\ref{fig:nfa_ablation} shows a set of comparisons.
Observing the grass area in red boxes, the result of variant 1 is dark and remaining thin haze. Variants 2 and 4 also have non-homogeneous fog residues in some areas.
% We can find that the full NFA can help dehazing network generate the best result compared to other variants.
%
Variant 3 generates obvious artifacts because forcing normalizing unaligned features to the same order of magnitude and adding them together makes the network difficult to train.
Only the full NFA achieves the best in brightness and haze removal.

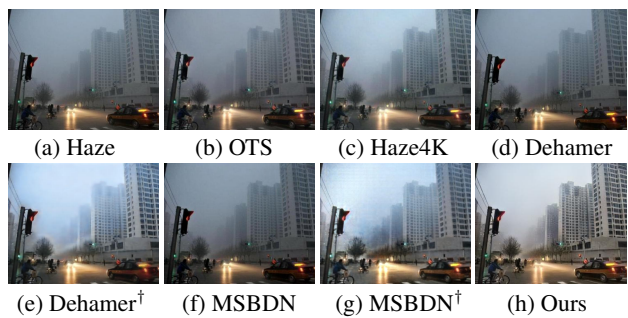
\begin{figure}[t]
    \centering
    \includegraphics[width=\linewidth]{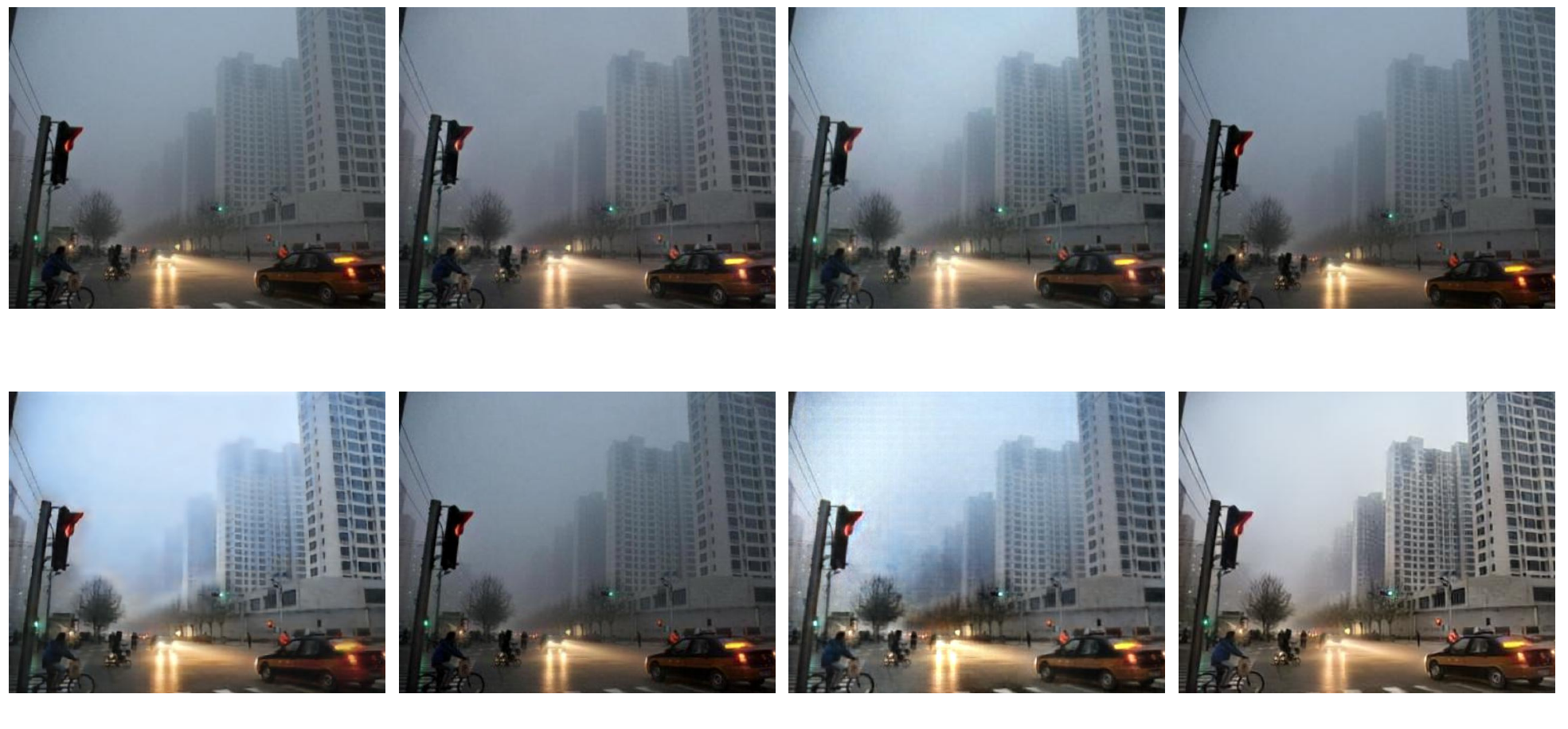}
    \put(-226, 54){\small{(a)~Haze}}
    \put(-165, 54){\small{(b)~OTS}}
    \put(-111, 54){\small{(c)~Haze4K}}
    \put(-53, 54){\small{(d)~Dehamer}}
    \put(-232.5, -5){\small{(e)~Dehamer$^\dag$}}
    \put(-170, -5){\small{(f)~MSBDN}}
    \put(-112, -5){\small{(g)~MSBDN$^\dag$}}
    \put(-47, -5){\small{(h)~Ours}}
    \vspace{-0.15cm}
    \caption{
        Ablation results of the proposed data generation pipeline.
        We retrain our dehazing network on OTS (b) and Haze4K (c) to verify the effectiveness of our generation pipeline.
        Dehamer and MSBDN are also retrained on our synthetic data, which are marked by $\dag$. 
        % \lichongyi{would be great if could show more obvious cases, say Ours can effectively remove all haze.}
    }
    \vspace{-0.55cm}
    \label{fig:data_ablation}
\end{figure}

\noindent\textbf{Effectiveness of the Phenomenological Degradation Pipeline.}
To prove that our proposed degradation pipeline for paired data generation can boost the capabilities of haze removal,
we retrain our RIDCP on two widely-used synthetic datasets,
which are OTS~\cite{li2019benchmarking} and Haze4K~\cite{zheng2021ultra}.
Notably, Haze4K is post-processed by DAD~\cite{shao2020domain}.
Moreover, we replace the training set from OTS with our synthetic data for transformer-based Dehamer and CNN-based MSBDN, thus demonstrating that it can generally bring gains.
The comparison results  are illustrated in Figure~\ref{fig:data_ablation}.
We can observe that our dehazing network cannot remove the haze under the training of OTS and Haze4K.
Besides, Dehamer and MSBDN can generate results with less haze and higher brightness with the help of our training data.
However, they still struggle in color recovery compared to our method, which also demonstrates the effectiveness of HQPs and our adaptation strategy.

\vspace{-0.3cm}
\section{Discussion}
\noindent\textbf{Conclusion.}
In this paper, we present a novel paradigm to revitalize deep dehazing networks towards the real world.
Our proposed phenomenological degradation pipeline synthesizes more realistic hazy data,
which achieves significant gains in haze removal.
Based on our observations and analysis, we introduce the high-quality priors in VQGAN to the dehazing network and progressively leverage their power, which finally builds our real image dehazing network via high-quality codebook priors (RIDCP).
Extensive experiments show the superiority of the proposed paradigm.

\noindent\textbf{Limitations and Future Work.}
In the process of doing our work on RIDCP, we observed that there are still some difficulties that are urgent to be addressed.
We leave the challenges here and hope that future work can address them
\begin{itemize}
    \vspace{-0.2cm}
    \item Existing dehazing methods including RIDCP can not process non-homogeneous haze well.
    \vspace{-0.2cm}
    \item Dehazing based on enhancement fashion is limited. Generative ability should be introduced for recovering extremely dense haze.
    \vspace{-0.2cm}
    \item We found that difficult to benchmark dehazing methods in quantitative fairly. Robust metrics for evaluating the quality of dehazing results are also needed.
\end{itemize}

%%%%%%%%% REFERENCES

\textbf{Acknowledgements.} This work is funded by the National Key Research and Development Program of China (NO.2018AAA0100400), Fundamental Research Funds for the Central Universities (Nankai University, NO.63223050), China Postdoctoral Science Foundation (NO.2021M701780). We are also sponsored by CAAI-Huawei MindSpore Open Fund.

\appendix
\twocolumn[{
% \renewcommand\twocolumn[1][]{#1}
% \maketitle
\begin{center}
    \captionsetup{type=figure}
    \includegraphics[width=\textwidth]{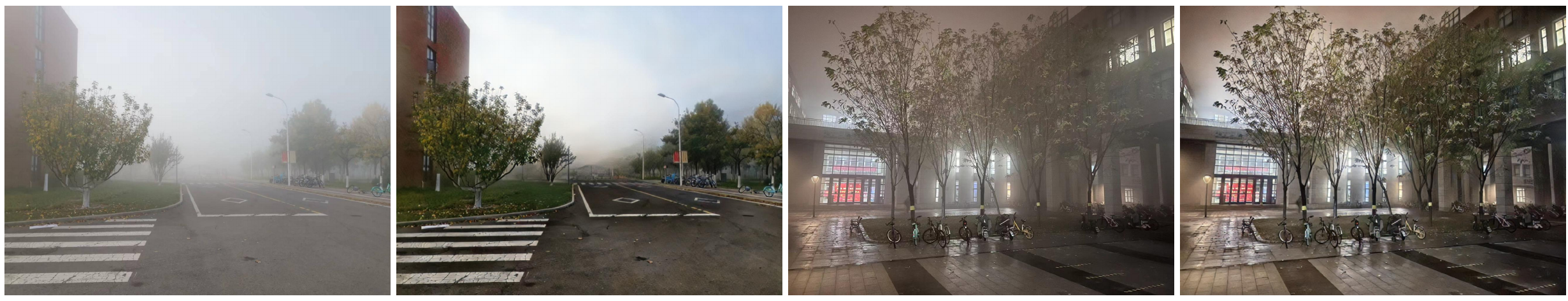}
    \captionof{figure}{Dehazing results of data captured by us. The proposed RIDCP performs well on both daytime and nighttime.}
    % \caption{Caption}
    \label{fig:teaser}
\end{center}
}]
\begin{appendixabstract}
Our supplementary materials give more details of our RIDCP and more experiments results, which can be summarized as follows:
\begin{itemize}
    \item We provide the detailed architectures and the training objectives of the pre-trained VQGAN network and our RIDCP.
    \item We provide more visual results on RTTS to demonstrate the superior performance of the proposed RIDCP.
    \item We provide more qualitative comparisons to prove the effectiveness of HQPs and the proposed phenomenological degradation pipeline.
    \item We provide a video demo to show our RIDCP's potential in real video dehazing.
\end{itemize}
\end{appendixabstract}

\section{Network Details}
\subsection{Detailed Architecture}
Table~\ref{tab:detail_ridcp} illustrates the detailed architecture of our RIDCP and the correspondence output size.
Each encoder layer is consist of a down-sampling convolutional layer with a sliding stride of $2$ and two residual layers~\cite{he2016deep}.
Each decoder layer is consist of an up-sampling operation, a convolutional layer, and two residual layers~\cite{he2016deep}.

\begin{table}[h]
    \centering
    \renewcommand\arraystretch{1.2}
    \begin{tabular}{p{1.8cm}<{\centering}|p{3.2cm}<{\centering}|p{1.8cm}<{\centering}}
        \hline \hline
         \textbf{Layers} & \textbf{Configurations} & \textbf{Output Size}\\ \hline
         Input & RGB Image & $h\times w \times 3$\\ \hline
         Conv1 & $c=64$ \quad $k=3$ & $h \times w \times 64$ \\ \hline
         Enc1 & $c=128$ \quad $k=3$ & $\frac{h}{2} \times \frac{w}{2} \times 128$ \\ \hline
         Enc2 & $c=256$ \quad $k=3$ & $\frac{h}{4} \times \frac{w}{4} \times 256$ \\ \hline
         RSTB & $\begin{bmatrix} c=256\\ h=8\\ ws=8 \end{bmatrix} \times 4$ & $\frac{h}{4} \times \frac{w}{4} \times 256$\\ \hline
         Conv2 & $c=512$ \quad $k=1$ & $\frac{h}{4} \times \frac{w}{4} \times 512$ \\ \hline
         Codebook & $c=512$ \quad $K=1024$ & $\frac{h}{4} \times \frac{w}{4} \times 512$ \\ \hline
         Conv3 & $c=256$ \quad $k=1$ & $\frac{h}{4} \times \frac{w}{4} \times 256$ \\ \hline
         Dec1\_vq & $c=128$ \quad $k=3$ & $\frac{h}{2} \times \frac{w}{2} \times 128$ \\ \hline
         Dec2\_vq & $c=64$ \quad $k=3$ & $h \times w \times 64$ \\ \hline
         Dec1 & $c=128$ \quad $k=3$ & $\frac{h}{2} \times \frac{w}{2} \times 128$ \\ \hline
         Dec2 & $c=64$ \quad $k=3$ & $h \times w \times 64$ \\ \hline
         Conv4 & $c=3$ \quad $k=3$ & $h \times w \times 3$ 
         \\ \hline \hline
    \end{tabular}
    \caption{Architecture details of the RIDCP. 
    $c$ denotes the output channel number, $k$ represents the kernel size, and $K$ is the codebook size.
    $h$ and $ws$ are number of heads and window size respectively.}
    \vspace{-0.5cm}
    \label{tab:detail_ridcp}
\end{table}

\subsection{Traning Objectives}
\textbf{VQGAN.}
% \wrq{I suggest putting this section in supplementary materials.}
Since the vector-quantized operation is non-differentiable,
VQGAN is end-to-end trained by copying the gradients of $\mathbf{G}_{vq}$ to $\mathbf{E}_{vq}$~\cite{bengio2013estimating}.
The optimization strategy can be divided into two parts, which are minimizing the loss after reconstruction and after feature matching, respectively.
For the first part, the loss function can be formulated as:
\begin{equation}
    \mathcal{L}_{rec} = ||x^\prime - x||_1 + \mathcal{L}_{per} + \mathcal{L}_{adv},
\end{equation}
where $\mathcal{L}_{per}$ and $\mathcal{L}_{adv}$ are perceptual loss~\cite{Johnson2016Perceptual} and adversarial loss~\cite{ledig2017photo}, respectively. And for codebook optimization, the loss function can be written as:
\begin{equation}
\begin{split}
    \mathcal{L}_{codebook} = ||sg(\hat{z}) - z^q||^2_2 + \beta ||sg(z^q) - \hat{z}||_2^2 \\+ \gamma||CONV(z^q) - \phi(x)||_2^2,
\end{split}
\end{equation}
where $sg(\cdot)$ is the stop-gradient operation, and $\beta=0.25, \gamma=0.1$ respectively. 
The last term of $\mathcal{L}_{codebook}$ is a semantic guided regularization term follow~\cite{chen2022real}, where $CONV$ is a simple convolutional layer, and $\phi$ is the pretrained VGG19~\cite{vgg}. Finally, the total loss of VQGAN is:
\begin{equation}
    \mathcal{L}_{vq} = \mathcal{L}_{rec} + \mathcal{L}_{codebook}.
\end{equation}

\textbf{RIDCP.}
For encoder $\mathbf{E}$, we use pretrained VQGAN to teach it to find the correct code.
Assuming that the input hazy image is $x_{h}$ and the clear counterpart is $x_{gt}$,
we can get features $\hat{z}_h=\mathbf{E}(x_h)$ and $z_{gt}^q=\mathcal{M}(\mathbf{E}_{vq}(x_{gt}))$.
The loss function $\mathcal{L}_{\mathbf{E}}$ to optimize $\mathbf{E}$ can be formulated as:
\begin{equation}
\begin{split}
    \mathcal{L}_{\mathbf{E}} = ||\hat{z}_h - z_{gt}^q||_2^2 + \lambda_{style}||\Psi (\hat{z}_h - \Psi(z_{gt}^q))||_2^2 + \\
    \lambda_{adv} \sum \limits_i -\mathbb{E}[D(\hat{z}_h^i)],
\end{split}
\end{equation}
where $\Psi$ is the Gram matrix calculation to build style loss~\cite{gondal2018unreasonable} and $D$ is the discriminator to supervise $\mathbf{E}$ adversarially. 
And $x_{gt}$ is used for supervising $\mathbf{G}$, which can be written as:
\begin{equation}
    \mathcal{L}_{\mathbf{G}} = ||y - x_{gt}||_1 + \lambda_{per}||\phi(y) - \phi(x_{gt})||_2^2,
\end{equation}
where $y$ is the output and $\phi$ the pretrained VGG16~\cite{journals/corr/SimonyanZ14a}.
Besides, the gradients of $\mathcal{L}_{\mathbf{G}}$ do not propagate backwards to $\mathbf{E}$.

\section{Experiments Results}
Since there is no ground-truth for real image dehazing tasks, quantitative metrics are difficult to reflect the true performance of the dehazing algorithms.
Meanwhile, we provide extensive qualitative results in the section to further demonstrate the superior performance of the proposed RIDCP and the effectiveness of each key component.

\subsection{More Visual Results}
\begin{figure*}
    \centering
    \vspace{-0.2cm}
    \includegraphics[width=\textwidth]{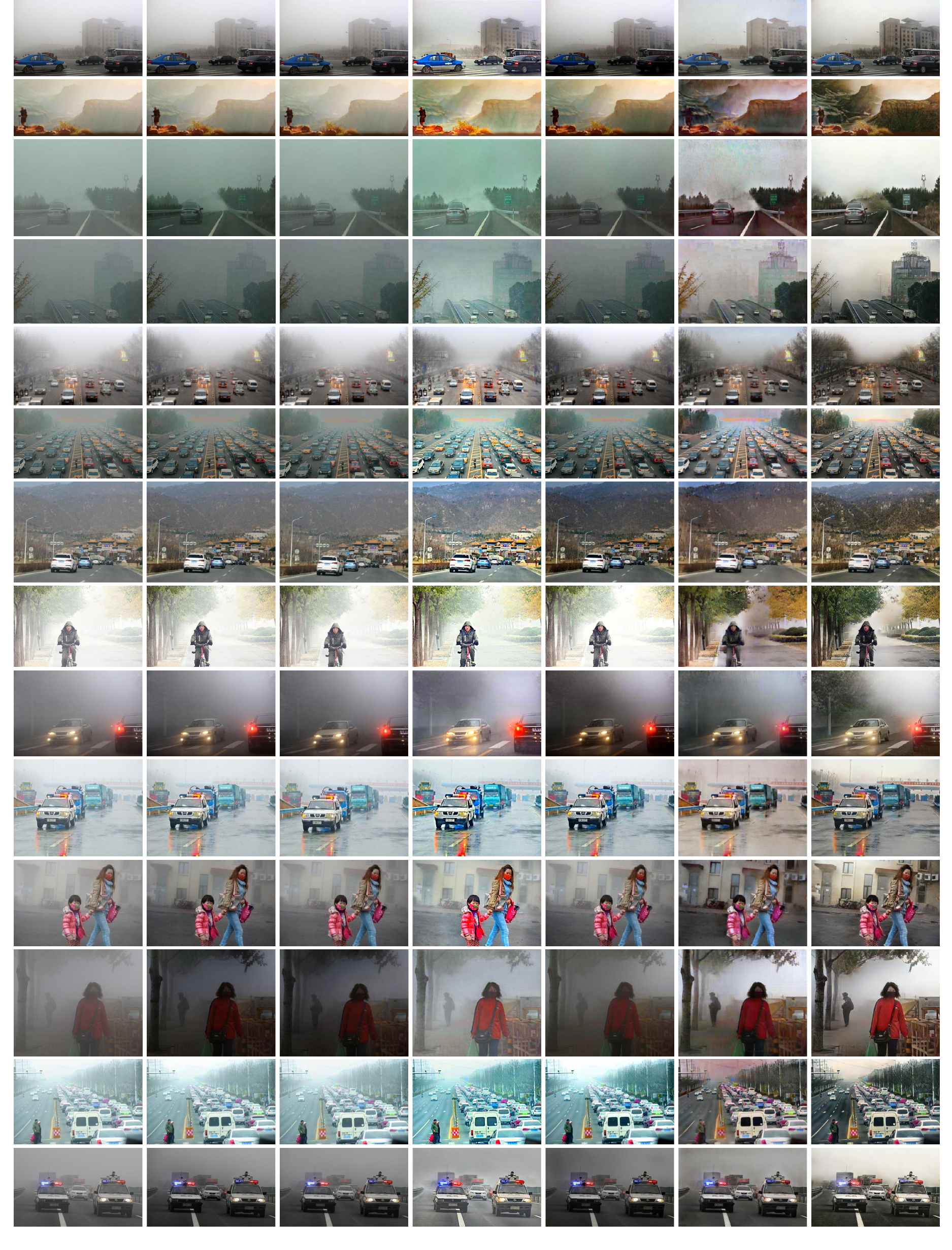}
    \put(-484.5, 0){\small{(a)~Hazy image}}
    \put(-417.5, 0){\small{(b)~MSBDN~\cite{dong2020multi}}}
    \put(-349, 0){\small{(c)~Dehamer~\cite{guo2022image}}}
    \put(-270, 0){\small{(d)~PSD~\cite{chen2021psd}}}
    \put(-199, 0){\small{(e)~D4~\cite{yang2022self}}}
    \put(-133, 0){\small{(f)~DAD~\cite{shao2020domain}}}
    \put(-59, 0){\small{(g)~RIDCP}}
    \vspace{-0.2cm}
    \caption{More visual comparisons on RTTS. \textbf{Zoom-in for best view.}}
    \label{fig:rtts_compare1}
\end{figure*}

\begin{figure*}
    \centering
    \includegraphics[width=\textwidth]{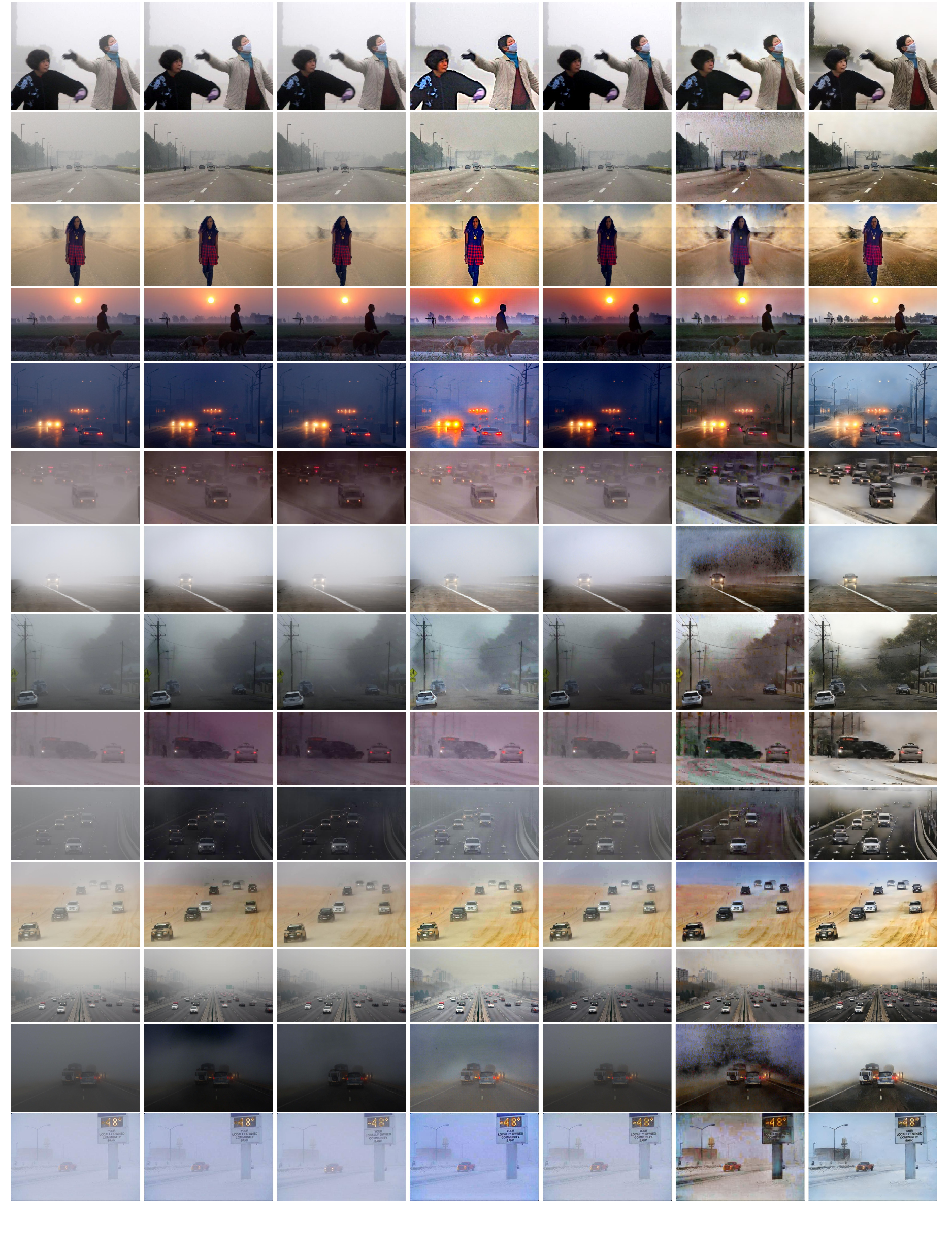}
    \put(-484.5, 11.5){\small{(a)~Hazy image}}
    \put(-417.5, 11.5){\small{(b)~MSBDN~\cite{dong2020multi}}}
    \put(-349, 11.5){\small{(c)~Dehamer~\cite{guo2022image}}}
    \put(-270, 11.5){\small{(d)~PSD~\cite{chen2021psd}}}
    \put(-199, 11.5){\small{(e)~D4~\cite{yang2022self}}}
    \put(-133, 11.5){\small{(f)~DAD~\cite{shao2020domain}}}
    \put(-59, 11.5){\small{(g)~RIDCP}}
    \vspace{-0.5cm}
    \caption{More visual comparisons on RTTS. \textbf{Zoom-in for best view.}}
    \label{fig:rtts_compare2}
\end{figure*}

\begin{figure*}
    \centering
    \includegraphics[width=\textwidth]{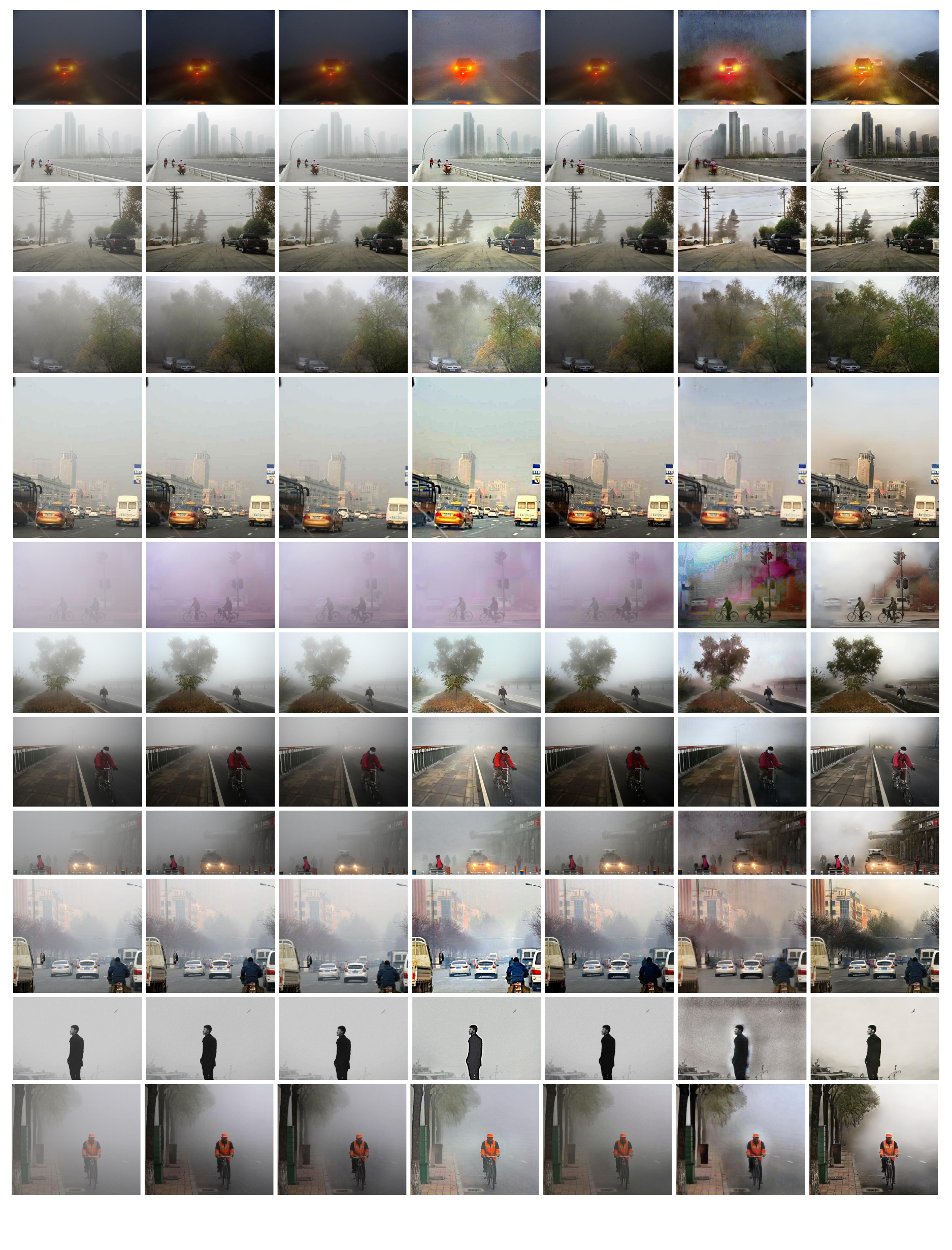}
    \put(-484.5, 11.5){\small{(a)~Hazy image}}
    \put(-417.5, 11.5){\small{(b)~MSBDN~\cite{dong2020multi}}}
    \put(-349, 11.5){\small{(c)~Dehamer~\cite{guo2022image}}}
    \put(-270, 11.5){\small{(d)~PSD~\cite{chen2021psd}}}
    \put(-199, 11.5){\small{(e)~D4~\cite{yang2022self}}}
    \put(-133, 11.5){\small{(f)~DAD~\cite{shao2020domain}}}
             \put(-59, 11.5){\small{(g)~RIDCP}}
    \vspace{-0.5cm}
    \caption{More visual comparisons on RTTS. \textbf{Zoom-in for best view.}}
    \label{fig:rtts_compare3}
\end{figure*}

Figure~\ref{fig:teaser} presents two dehazing cases on our own-captured data.
Our dehazing method performs well on both daytime and nighttime scenes.
Figure~\ref{fig:rtts_compare1}, \ref{fig:rtts_compare2} and \ref{fig:rtts_compare3} illustrate more visual comparisons with several state-of-the-art methods on RTTS~\cite{li2019benchmarking} dataset.
As we can see, the proposed RIDCP achieves satisfactory performance and maintains stable dehazing ability in scenes with dense haze and heavy color bias.

\subsection{Ablation Study}
\begin{figure}
    \centering
    \includegraphics[width=\linewidth]{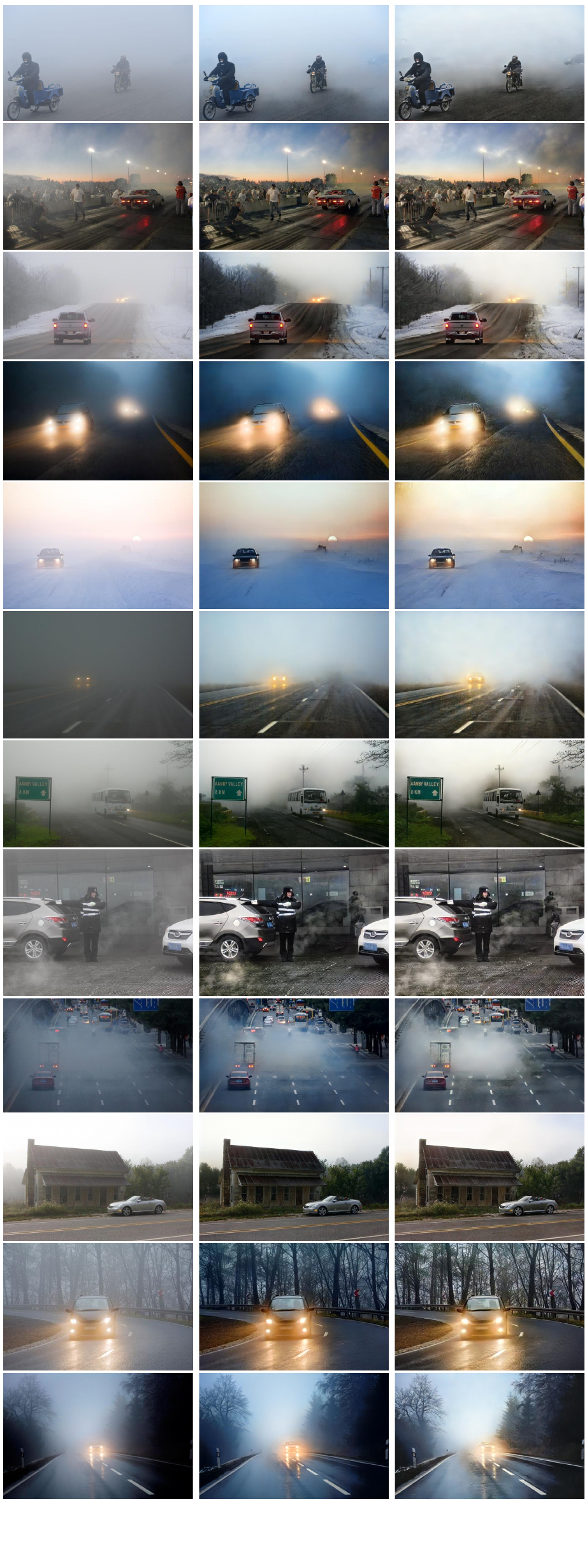}
    \put(-225, 16){\small{(a)~Hazy image}}
    \put(-145, 16){\small{(b)~w/o HQPs}}
    \put(-68, 16){\small{(c)~Full model}}
    \vspace{-0.5cm}
    \caption{Ablation results on HQPs.}
    \label{fig:hqps_compare}
\end{figure}

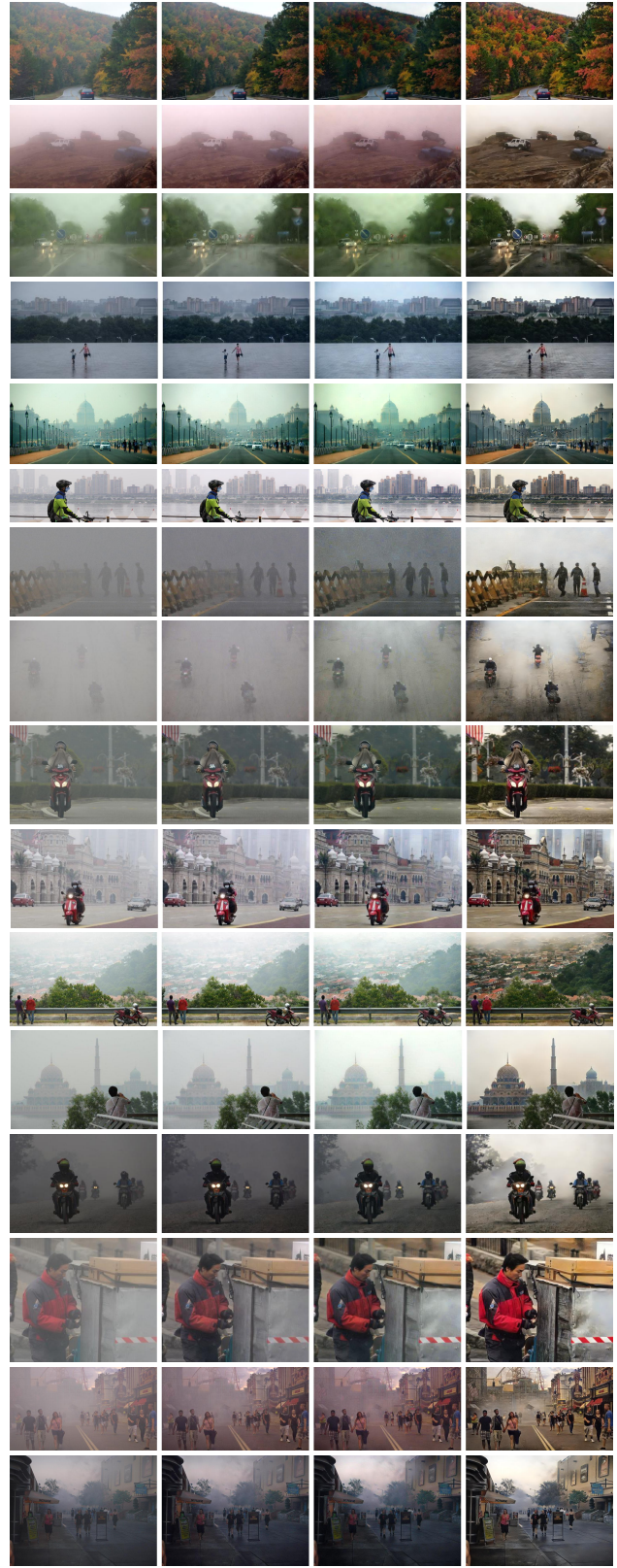
\begin{figure}
    \centering
    \includegraphics[width=\linewidth]{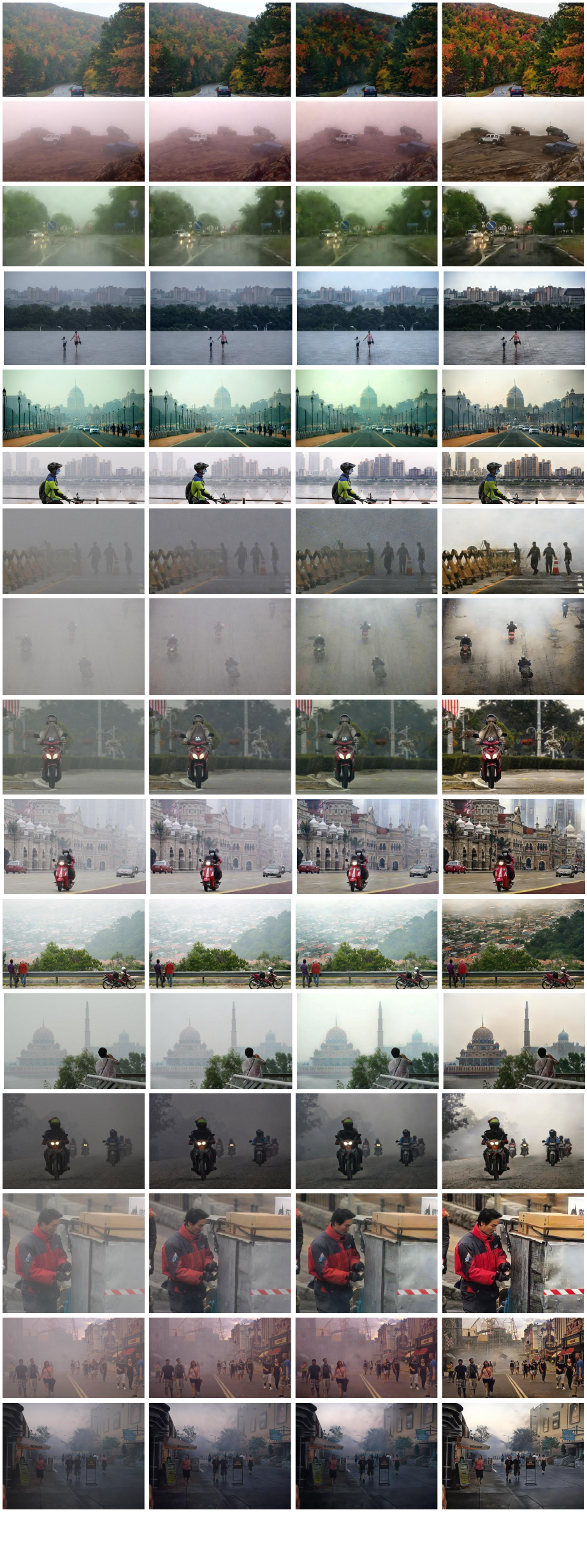}
    \put(-237, 10){\small{(a)~Hazy image}}
    \put(-165, 10){\small{(b)~OTS}}
    \put(-110, 10){\small{(c)~Haze4K}}
    \put(-57, 10){\small{(d)~Our pipeline}}
    \vspace{-0.5cm}
    \caption{Ablation results on the proposed phenomenological degradation pipeline.}
    \label{fig:data_compare}
\end{figure}

We analyze the effectiveness of HQPs and the proposed phenomenological degradation pipeline.
In Figure~\ref{fig:hqps_compare}, we can observe that HQPs can help the network generate results with better brightness and lower color bias.
Figure~\ref{fig:data_compare} shows the significant improvement in dehazing capability brought by our pipeline.

\section{Broader Impacts}
Our RIDCP performs well on real-world hazy scenes, which can possibly be applied to some industrial tasks like automatic driving and computational photography.
Moreover, the proposed phenomenological degradation pipeline can also generally boost the performance of dehazing algorithms, 
which is beneficial for the development of real image dehazing. 
Thus, we believe our work will bring positive impacts on both academia and industry.
As a typical low-level vision work, this paper will not bring negative impacts to society.

%%%%%%%%% REFERENCES

% \begin{figure}
%     \centering
%     \includegraphics[width=\linewidth]{figures/supp/no_chm.pdf}
%     \put(-225, 16){\small{(a)~Hazy image}}
%     \put(-145, 16){\small{(b)~w/o CHM}}
%     \put(-68, 16){\small{(c)~Full model}}
%     \vspace{-0.5cm}
%     \caption{Caption}
%     \label{fig:adjust_compare}
% \end{figure}

\clearpage
{\small
\bibliographystyle{ieee_fullname}
\bibliography{egbib}
}

\end{document}